\documentclass[11pt]{article}

\usepackage{mathrsfs,amsmath,amsfonts,amssymb,amstext,amscd,bm,bbm,dsfont,pifont, 
            amsthm,stmaryrd,euscript,color,xcolor,accents,xr} %bm,bbm, dsfont,pifont,

\usepackage[backref]{hyperref}

\usepackage{algorithmic}
\usepackage{algorithm}
\usepackage{url}
\usepackage{mathtools}
\usepackage{epsfig}
\usepackage{float}
\usepackage{appendix}
\usepackage{amstext}
\usepackage{floatflt}
\usepackage{nicefrac}
\usepackage{amsmath}
\usepackage{anysize,hyperref}
\usepackage{enumerate}
\usepackage{epsfig}
\usepackage{hyperref}
\usepackage{graphicx}
\numberwithin{equation}{section} 

\usepackage{mathtools}
\usepackage[backref]{hyperref}

\input{macros_nicole}
\newcommand{\ols}{{\rm RR}}%{{\rm OLS}}
\newcommand{\olsest}{ \overline{w}^{{ \small{ \rm RR}}}}%{ \overline{w}^{{ \small{ \rm OLS}}}}

\newcommand{\sgd}{{\tiny{\rm SGD}}}%{{\rm OLS}}

\newcommand{\bias}{{\rm Bias}}
\newcommand{\var}{{\rm Var}}

\newcommand{\w}{\overline{\overline w}}
\newcommand{\bb}{\overline{\overline b}}
\newcommand{\vv}{\overline{\overline v}}

\newcommand{\citet}{\cite}

\usepackage[linecolor=magenta!60!, backgroundcolor=magenta!10!,textwidth=1.6cm, textcolor=magenta]{todonotes}

\title{Local SGD in Overparameterized Linear Regression}

\author{Mike Nguyen \\
Technical University  of Braunschweig \\
\texttt{mike.nguyen@tu-braunschweig.de} 
\and
Charly Kirst \\
Technical University  of Braunschweig  \\ 
\texttt{c.kirst@tu-braunschweig.de}
\and
Nicole M\"ucke\\
Technical University  of Braunschweig  \\ 
\texttt{nicole.muecke@tu-braunschweig.de}

}
\date{\today}

\begin{document}

\maketitle

\begin{abstract}
\noindent
We consider distributed learning using constant stepsize SGD (DSGD) over several devices, each sending a final model update to a central server. 
In a final step, the local estimates are aggregated. We prove in the setting of overparameterized linear regression general upper bounds 
with matching lower bounds and derive learning rates for specific data generating 
distributions. We show that the excess risk is of order of the variance  provided the number of local nodes grows not too large with the 
global sample size. 
We further compare the sample complexity of DSGD with the sample complexity of distributed ridge regression (DRR) and show that 
the excess SGD-risk is smaller than the excess RR-risk, where both sample complexities are of the same order.  
\end{abstract}

\section{INTRODUCTION}

Deep neural networks possess powerful generalization properties in various machine learning applications,
despite being overparameterized. It is generally believed that the optimization algorithm itself,
e.g., stochastic gradient descent (SGD), implicitly regularizes such overparameterized models. % \todo{[29]}. 
%here, (unregularized) overparameterized models could admit numerous global and local minima
%(many of which generalize poorly [29, 21]), yet SGD tends to find solutions that generalize well, even
%in the absence of explicit regularizers [22, 29, 19]. 
This regularizing effect due to the choice of the 
optimization algorithm is often referred to as \emph{implicit regularization}. % \todo{[22]}. 
A refined understanding of this phenomenon 
was recently gained in the setting of linear regression (to be considered as a reasonable approximation of neural network learning) 
for different variants of SGD. 
Constant stepsize SGD (with last iterate or tail-averaging)  
is investigated in \cite{jain2018markov}, in \cite{DieuBa16} in an RKHS frameowrk and also in \cite{mucke2019beating} with additional mini-batching, 
see also \cite{mucke2020stochastic} for a more general analysis in Hilbert scales. In  
\cite{zou2021benign, zou2021benefits} it is shown that benign overfitting also occurs for SGD. 
Multi-pass SGD is analyzed in \cite{LinCamRos16, Kakade16, LinRos17, zou2022risk} while last iterate bounds 
can be found in \cite{jain2019making, wu2022last, varre2021last}.

Despite the attractive statistical properties of all these SGD variants,
the complexity of computing regression estimates prevents it from being routinely used in
large-scale problems. More precisely, the time complexity and space complexity of SGD and other regularization methods 
in a standard implementation scale as $\cO(n^\alpha)$, $\alpha \in [2,3]$. Such scalings are
prohibitive when the sample size $n$ is large.

Distributed learning (DL) based on a divided-and-conquer approach is an effective way to
analyze large scale data that can not be handled by a single machine. In this paper
we study a distributed learning strategy in linear regression (including
both underparameterized and overparameterized regimes) via (tail-) averaged stochastic 
gradient descent with constant stepsize (DSGD). The approach is quite simple and communication efficient: 
The training data is distributed across several computing nodes where on each a local SGD is run. In a final step, these local estimates are 
aggregated (a.k.a. \emph{one-shot SGD}). Local SGD has become state of the art in large scale distributed learning, 
showing a linear speed-up in the number of workers for 
convex problems, 
see e.g. \cite{mcdonald2009efficient, zinkevich2010parallelized, dieuleveut2019communication, stich2018local, spiridonoff2021communication} and references therein.  
  
The field of DL has gained increasing attention in statistical learning theory with the aim of deriving conditions under which 
minimax optimal rates of convergence can be guaranteed, 
see e.g. \citet{chen2014split}, \citet{mackey2011divide}, \citet{xu2019distributed}, \citet{fan2019distributed}, \citet{shi2018massive}, \citet{battey2018distributed}, 
\citet{fan2021communication}, \citet{bao2021one}. 
%We give a more detailed overview over 
%approaches that are most closely related to our approach. For a general overview we refer to \citet{bekkerman2011scaling} 
%and the recent review \citet{gao2021review}.  
Indeed, the learning properties of DL in regression settings over Hilbert spaces are widely well understood. 
The authors in \citet{zhang2015divide} analyze distributed (kernel) ridge regression and 
show optimal learning rates with appropriate regularization, 
provided the number of machines increases sufficiently slowly with the sample size, though under restrictive assumptions on the eigenfunctions of the kernel 
integral operator. This has been alleviated in \citet{lin2017distributed}. However, in these works the number of machines \emph{saturates} 
if the target is very smooth, meaning that large parallelization seems not possible in this regime. 
%This is somewhat counterintuitive as smooth signals are easier to reconstruct.      
%To overcome this issue, the authors \citet{chang2017distributed} utilize additional unlabeled data, leading to a slight improvement. 

An extension of these works to more general spectral regularization algorithms  for nonparametric least square regression in 
(reproducing kernel) Hilbert spaces is given in \citet{guo2017learning}, \citet{mucke2018parallelizing}, including gradient descent (\cite{lin2018distributed}) 
and stochastic gradient descent (\cite{lin2018optimal}). The recent work  \cite{tong2021distributed} studies DL for functional linear regression. 

We finally mention the work of \cite{mucke2022data}, where distributed ordinary least squares (DOLS) in overparameterized linear regression is studied, i.e. 
one-shot OLS without any explicit or implicit regularization. It is shown that the number of workers acts as a regularization parameter itself.

\paragraph{Contributions.} 
We analyze the performance of DSGD with constant stepsize in overparameterized linear regression and provide upper bounds with matching lower bounds 
for the excess risk under suitable noise assumptions. Our results show that optimal rates of convergence can be achieved if the number of local nodes 
grows sufficiently slowly with the sample size. The excess risk as a function of data splits remains constant until a certain threshold is reached. This threshold depends 
on the structural assumptions imposed on the problem, i.e. on the eigenvalue decay of the Hessian and the coefficients of the true regression parameter.

We additionally perform a comparison between DSGD and DRR, showing that the excess risk of DSGD is upper bounded by the excess risk of DRR under an assumption 
on the sample complexity (SC) of DSGD, depending on the same structural assumptions. We show that the SC of DSGD remains within constant factors of the SC 
of DRR.

Our analysis extends known results in this direction from \cite{zou2021benign, zou2021benefits} for the single machine case to the distributed learning 
setting and from DOLS in \cite{mucke2022data} to 
SGD with implicit regularization.

\paragraph{Organization.}
In Section \ref{sec:setup} we define the mathematical
framework needed to present our main results in Section \ref{sec:main-results},  where we provide a theoretical analysis of DSGD with a discussion of our results. 
In Section \ref{sec:benefit} we compare DSGD with DRR while Section \ref{sec:numerics} is devoted to showing some numerical illustrations.  
The proofs a deferred to the Appendix.

\paragraph{Notation.} 
By $\cL(\cH_1, \cH_2)$ we denote the space of bounded linear operators between real Hilbert spaces $\cH_1$, $\cH_2$. 
We write $\cL(\cH, \cH) = \cL(\cH)$. For $\bA \in \cL(\cH)$ we denote by $\bA^T$ the adjoint operator. 
By $\bA^\dagger$ we denote the pseudoinverse of $\bA$ and for $w \in \cH$ we write $||w||^2_{\bA} := ||\bA^{\frac{1}{2}} w||$ for an PSD operator $\bA$. 

We let $[n]=\{1,...,n\}$ for every $n \in \mbn$.  For two positive sequences $(a_n)_n$, $(b_n)_n$ we write $a_n \lesssim b_n$ if $a_n \leq c b_n$ 
for some $c>0$ and $a_n \simeq  b_n$ if both $a_n \lesssim b_n$ and $b_n \lesssim a_n$.

%%%%%%%%%%%%%%%%%%%%%%%%%%%%%%%%%%%%%%%%%%%%%%%%%%%%%%%%%%%%%%%%%%%%%%%%%%%%%%%%%%%%%%%%%%%%%%%%%%%%%%%%%%%%
%%%%%%%%%   Problem Setting     %%%%%%%%%%%%%%%%%%%%%%%%%%%%%%%%%%%%%%%%%%%%%%%%%%%%%%%%%%%%%%%%%%%%%%%%%%%%%%%%%%%%%%
%%%%%%%%%%%%%%%%%%%%%%%%%%%%%%%%%%%%%%%%%%%%%%%%%%%%%%%%%%%%%%%%%%%%%%%%%%%%%%%%%%%%%%%%%%%%%%%%%%%%%%%%%%%%

\section{SETUP}
\label{sec:setup}

In this section we provide the mathematical framework for our analysis. More specifically, we 
introduce distributed SGD and state the main assumptions on our model.

%%%%%%%%%%%%%%%%%%%%%%%%%%%%%%%%%%%%%%%%%%%%%%%%%%%%%%%%%%%%%%%%%%%%%%%%%%%
%%%%%%%%%%%%%%%%%%%%%%%%%%%%%%%%%%%%%%%%%%%%%%%%%%%%%%%%%%%%%%%%%%%%%%%%%%%

\subsection{SGD and linear regression}

We consider a linear regression model over a real separable Hilbert space $\cH$ in random design. More precisely, 
we are given a random covariate vector $x \in \cH$ and a random output $y \in \mbr$ following the model  
\begin{equation}
\label{eq:model}  
y = \inner{w^*, x} + \epsilon \;, 
\end{equation}
where $\epsilon \in \mbr$ is a noise variable. We will impose some assumptions on the noise model in Section \ref{sec:main-results}. 
The true regression parameter $w^* \in \cH$ minimizes the least squares test risk, i.e.  
\[  L(w^*) = \min_{w \in \cH} L(w)\;, \quad L(w) := \frac{1}{2}\mbe[(y - \inner{w , x})^2] \;, \]
where the expectation is taken with respect to the joint distribution $\mbp$ of the pair $(x,y) \in \cH \times \mbr$. More specifically, 
we let $w^*$ be the minimum norm element in the set of all minimizers of $L$.

%\note{distribution $(x,y)\sim \mbp$, marginal of $x$ is $\mbp_x$}

To derive an estimator $\hat w \in \cH$ for $w^*$ we are given an i.i.d. dataset 
\[   D:= \{ (x_1, y_1), ..., (x_n, y_n) \} \subset  \cH \times \mbr \;, \] 
following the above model \eqref{eq:model}, i.e., 
\[  \bY = \bX w^* + \beps \;, \]
with i.i.d. noise 
$\beps = (\eps_1, ..., \eps_n) \in \mbr^n$. 
The corresponding random vector of outputs is denoted as $\bY=(y_1,\ldots, y_n)^T \in \mbr^n$ 
and we arrange the data $x_j \in \cH$ into a {\em data matrix} 
$\bX \in \cL(\cH, \mbr^n)$ by setting $(\bX v)_j =\langle x_j,v \rangle$ for $v \in \cH, 1 \leq j \leq n$. 
If $\cH=\mbr^d$, then $\bX$ is a $n \times d$ matrix (with row vectors $x_j$). 
We are particular interested in the overparameterized regime, i.e. where $dim(\cH) > n$.

In the classical setting of stochastic approximation with constant stepsize, the SGD iterates are computed by the recursion 
\begin{equation*}
w_{t+1} = w_t - \gamma ( \inner{w_t , x_t} - y_t  ) x_t \;, \;\; t=1, ..., n \;, 
\end{equation*} 
with some initialization $w_1 \in \cH$ and where $\gamma > 0$ is the stepsize. The tail average of the iterates is denoted by 
\begin{equation}
\label{eq:ave}
  \bar w_{\frac{n}{2}: n} := \frac{1}{n-n/2} \sum_{t=n/2+1}^n w_t \;,   
\end{equation}
and where we denote by $\bar w_{n}:=\bar w_{0: n}$ the full (uniform) average.

Various forms of SGD (with iterate averaging, tail averaging, multi passes) 
in the setting of overparameterized linear regression has been analyzed recently in \cite{zou2021benign}, 
%\note{some notes about SGD and benign overfitting papers ... averaging \cite{zou2021benign} 
\cite{wu2022last}, \cite{zou2022risk}, respectively. In particular, the phenomenon of \emph{benign overfitting} 
is theoretically investigated in these works. It could be shown that benign overfitting occurs in this setting, 
i.e. the SGD estimator fits training data very well and still generalizes. 

We are interested in this phenomenon for localized SGD, i.e. when our training data is distributed over several computing devices.  
%implicit regularization benefits \cite{zou2021benefits} }

%%%%%%%%%%%%%%%%%%%%%%%%%%%%%%%%%%%%%%%%%%%%%%%%%%%%%%%%%%%%%%%%%%%%%%%%%%%
%%%%%%%%%%%%%%%%%%%%%%%%%%%%%%%%%%%%%%%%%%%%%%%%%%%%%%%%%%%%%%%%%%%%%%%%%%%

\subsection{Local SGD }

In the distributed setting, our data are evenly divided into $M \in \mbn$ local disjoint subsets  
\[ D = D_1 \cup ... \cup D_M   \] 
of size $|D_m|=\frac{n}{M}$, for $m=1,...,M$. To each local dataset we associate 
a \emph{local design matrix} $\bX_m \in \cL(\cH,\mbr^{\frac{n}{M}})$ (build with local row vectors $x^{(m)}_j$) with local output vector
$\bY_m \in \mbr^{\frac{n}{M}}$ and a local noise vector $\beps_m \in \mbr^{\frac{n}{M}}$.

The local SGD iterates  are defined as  
\begin{equation*}
w^{(m)}_{t+1} = w^{(m)}_t - \gamma \left( \inner{w^{(m)}_t , x^{(m)}_t} - y_t  \right) x^{(m)}_t \;, 
\end{equation*} 
for $ t=1, ..., \frac{n}{M}$ and $m=1,...,M$. The averaged local iterates $\bar w^{(m)}_{\frac{n}{M}}$ are computed according to \eqref{eq:ave}. 
We are finally interested in the uniform average of the local SGD iterates, building a global estimator: 
\begin{equation*}
\w_M := \frac{1}{M} \sum_{m=1}^M  \bar w^{(m)}_{\frac{n}{M}} \;. 
\end{equation*}

Distributed learning in overparameterized linear regression is studied in \cite{mucke2022data} for the 
ordinary least squares estimator (OLS), i.e. without any implicit or explicit regularization and with local 
interpolation. It is shown that local overfitting is harmless and regularization is done by the number of data splits.

We aim at finding optimal bounds for the excess risk 
\[ \mbe\left[L (\w_M) \right] - L(w^*)  \;, \]
of distributed SGD (DSGD) with potential local overparameterization and  as function of the number of local nodes 
$M$ and under various model assumptions, to be given in the next section.

%%%%%%%%%%%%%%%%%%%%%%%%%%%%%%%%%%%%%%%%%%%%%%%%%%%%%%%%%%%%%%%%%%%%%%%%%%%%%%%%%%%%%%%%%%%%%%%%%%%%%%%%%%%%
%%%%%%%%%   MAIN RESULTS     %%%%%%%%%%%%%%%%%%%%%%%%%%%%%%%%%%%%%%%%%%%%%%%%%%%%%%%%%%%%%%%%%%%%%%%%%%%%%%%%%%%%%%
%%%%%%%%%%%%%%%%%%%%%%%%%%%%%%%%%%%%%%%%%%%%%%%%%%%%%%%%%%%%%%%%%%%%%%%%%%%%%%%%%%%%%%%%%%%%%%%%%%%%%%%%%%%%

\section{MAIN RESULTS}
\label{sec:main-results}

In this section we present our main results. To do so, we first impose some model assumptions.

\vspace{0.2cm}

\begin{definition}
\begin{enumerate}
\item
We define the second moment of $x \sim \mbp_x$ to be the operator $\bH: \cH \to \cH$, given by 
\[  \bH := \mbe[ x \otimes x ] = \mbe[ \inner{\cdot , x} x] \;.\]
\item
The fourth moment operator $\bM: \cL(\cH) \to \cL(\cH)$ is defined by 
\[ \bM := \mbe[ x \otimes x\otimes x \otimes x ] \;,\]
with $\bM (\bA) (w) = \mbe[  \inner{x , \bA x} \inner{w , x} x ] $, for all $w \in \cH$. 
\item
The covariance operator of the gradient noise at $w^*$ is defined as $\bbS : \cH \to \cH$, 
\[ \bbS := \mbe[ (\inner{w^* , x} - y)^2\; x \otimes x] \;. \]
\end{enumerate}
\end{definition}

\vspace{0.2cm}

\begin{assumption}[Second Moment Condition]
\label{ass:second-moment}
We assume that $\mbe[y^2 | x] < \infty$ almost surely. 
Moreover, we assume that the trace of $\bH$ is finite, i.e., $\tr[\bH] < \infty$. 
\end{assumption}

\vspace{0.2cm}

\begin{assumption}[Fourth Moment Condition]
\label{ass:fourth-moment}
We assume there exists a positive constant $\tau >0$ such that for any PSD operator $\bA$, we have 
%\[  \mbe[ x \otimes x \bA x \otimes x] \preceq \tau \tr[\bH \bA] \bH \;. \]
\[ \bM(\bA ) \preceq \tau \tr[\bH \bA] \bH \;. \]
\end{assumption}

Note that this assumption holds if $\bH^{-1}x$ is sub-Gaussian, being a standard assumption in 
least squares regression, see e.g. \cite{bartlett2020benign}, \cite{zou2021benign}, \cite{tsigler2020benign}.

\begin{assumption}[Noise Condition]
\label{ass:noise}
Assume that
\[ \sigma^2 := ||\bH^{-\frac{1}{2}} \bbS \bH^{-\frac{1}{2}} || < \infty \;. \]
\end{assumption}

This assumption on the noise is standard in the literature about averaged SGD, see e.g. 
\cite{zou2021benign}, \cite{DieuBa16}.

We introduce some further notation involving the second moment operator $\bH$: We denote the 
eigendecomposition as 
\[  \bH = \sum_{j = 1}^{\infty} \lam_j v_j \otimes v_j \;, \]
where the $\lam_1 \geq \lam_2 \geq ... $ are the eigenvalues of $\bH$ and the $v_j's$ are the corresponding eigenvectors. 
For $k \geq 1$, we let 
\[ \bH_{0:k} := \sum_{j=1}^k  \lam_j v_j \otimes v_j   \; , \;\;  \bH_{k:\infty} := \sum_{j=k+1}^\infty  \lam_j v_j \otimes v_j \;.\]
Similarly, 
\[ \bI_{0:k} = \sum_{j=1}^k   v_j \otimes v_j \; , \;\; \bI_{k:\infty} := \sum_{j=k+1}^\infty  v_j \otimes v_j \;.\]
A short calculation shows that for all $w \in \cH$ we have 
\[ ||w||^2_{\bH^\dagger_{0:k}} = \sum_{j=1}^k \frac{\inner{w, v_j}^2}{\lam_j} \,, \] 
\[    ||w||^2_{\bH_{k:\infty}}    = \sum_{j=k+1}^\infty \lam_j \inner{w, v_j}^2 \;. \]

\vspace{0.2cm}

We finally set 
\begin{equation}
\label{eq:Vstar}
  V_k(n,M) := \frac{k}{n} + \gamma^2 \frac{n}{M^2} \sum_{j=k+1}^\infty \lam_j^2 \;. 
\end{equation}

%%%%%%%%%%%%%%%%%%%%%%%%%%%%%%%%%%%%%%%%%%%%%%%%%%%%%%%%%%%%%%%%%%%%%%%%%
%%%%%%%%%%%%%%%%%%%%%%%%%%%%%%%%%%%%%%%%%%%%%%%%%%%%%%%%%%%%%%%%%%%%%%%%%

%\paragraph{Upper Bound.}

\subsection{Upper Bound}

We now present an upper bound for the averaged local SGD iterates. The proof relies on a bias-variance decomposition and  
is given in Appendix \ref{supp:proofsA}.

\vspace{0.2cm}

\begin{theorem}[DSGD Upper Bound]
\label{theo-main}
Suppose Assumptions \ref{ass:second-moment}, \ref{ass:fourth-moment} and \ref{ass:noise} are satisfied and 
let $\gamma < \frac{1}{\tau \tr[\bH]}$, $w_1=0$. The excess risk for the averaged local SGD estimate satisfies 
\[ \mbe\left[ L(\w_M) \right] - L(w^*)  \; \leq \; 2\bias (\w_M) \; + \;  2\var(\w_M)  \;, \]
where
\begin{align*}
  \bias (\w_M)  &\leq 
 \frac{M^2}{\gamma^2 n^2}  ||w^*||^2_{\bH^\dagger_{0:k^*}}+ ||w^*||^2_{\bH_{k^*:\infty}}\\
 &\; + \;  \frac{2\tau M^2\left(  ||w^*||^2_{\bI_{0:k^*}} +
 \gamma \frac{n}{M} || w^*||^2_{\bH_{k^*:\infty}} \right) }{\gamma  n (1-\gamma \tau \tr[\bH ]) } \cdot  V_{k^*}(n,M) 
\end{align*}
and 
\begin{align*}
\var(\w_M)  \; \leq \;   \frac{\sigma^2}{1-\gamma \tau \tr[\bH]} \cdot  V_{k^*}(n,M) \;,
\end{align*}
with $k^*=\max\{ k:\lam_k \geq \frac{M}{\gamma n} \}$.
\end{theorem}

The excess risk is upper bounded in terms of the bias and variance. Both terms crucially depend 
on the \emph{effective dimension} $k^*=\max\{ k:\lam_k \geq \frac{M}{\gamma n} \}$,  dividing the full Hilbert space $\cH$ into two parts. 
On the part associated to the first largest $k^*$  eigenvalues, the bias may decay faster than on the remaining tail part that is associated to the smaller eigenvalues,  
see \cite{zou2021benign} 
in the context of single machine SGD,  
\cite{bartlett2020benign, tsigler2020benign}, %\cite{tsigler2020benign} 
in the context of single machine ridge regression 
and \cite{mucke2022data} for distributed ordinary least squares. 

Our Theorem \ref{theo-main} reveals that the excess risk converges to zero if 
\[ ||w^*||^2_{\bH_{k^*:\infty}} \to 0\;,  \quad \frac{2M^2}{\gamma^2 n^2}  ||w^*||^2_{\bH^\dagger_{0:k^*}}\to 0 \]
%\[  \frac{2M^2}{\gamma^2 n^2}  ||w^*||^2_{\bH^\dagger_{0:k^*}}\to 0 \]
and $ V_{k^*}(n,M)   \to 0$ 
%\[   V_{k^*}(n,M)   \to 0 \]
as $n \to \infty$. This requires the eigenvalues of $\bH$ to decay sufficiently 
fast and to choose the number of local nodes $M=M_n$ to be a sequence of $n$. 
Note that we have to naturally assume $M_n \lesssim n$. 
In Subsection \ref{subsec:rates} we provide two specific examples of data distributions with specific choices for $(M_n)_{n \in \mbn}$ such that the above  
conditions are met, granting not only convergence but also 
providing explicit rates of convergence.

%%%%%%%%%%%%%%%%%%%%%%%%%%%%%%%%%%%%%%%%%%%%%%%%%%%%%%%%%%%%%%%%%%%%%%%%%
%%%%%%%%% LOWER BOUND
%%%%%%%%%%%%%%%%%%%%%%%%%%%%%%%%%%%%%%%%%%%%%%%%%%%%%%%%%%%%%%%%%%%%%%%%%

\subsection{Lower Bound}

Before we state the lower bounds for the excess risk of the DSGD estimator we need to impose some assumptions.

\begin{assumption}[Fourth Moment Lower Bound]
\label{ass:moment-lower}
We assume there exists a positive constant $\theta  >0$ such that for any PSD operator $\bA$, we have 
%\[  \mbe[ x \otimes x \bA x \otimes x] - \bH \bA \bH \succeq  \theta  \tr[\bH \bA] \bH \;. \]
\[  \bM(\bA )  - \bH \bA \bH \succeq  \theta  \tr[\bH \bA] \bH \;. \]

\end{assumption}

\vspace{0.2cm}

\begin{assumption}[Well-Specified Noise]
\label{ass:well}
The second moment operator $\bH$ is strictly positive definite with $\tr[\bH] < \infty$. 
Moreover, the noise $\epsilon$ in \eqref{eq:model} is independent of $x$ and satisfies 
%\[ \mbe[ \epsilon ] = 0 \;, \quad \mbe[\epsilon^2 ] = \sigma^2 \;. \]
\[ \epsilon \sim \cN(0 , \sigma^2_{noise}) \;. \]
\end{assumption}

\vspace{0.2cm}

We now come to the main result whose proof can be found in Appendix \ref{supp:proofsB}. 
\vspace{0.2cm}

\begin{theorem}[DSGD Lower Bound]
\label{theo:lower}
Suppose Assumptions \ref{ass:moment-lower} and \ref{ass:well} are satisfied. Assume $w_1=0$. The excess risk of the DSGD estimator 
satisfies
\begin{align*} 
 \mbe\left[ L(\w_M) \right] - L(w^*) &\geq  
 \frac{M(M-1)}{100 \gamma^2 n^2}
\left(  || w^*||^2_{\bH^\dagger_{0:k^*}} +   \frac{\gamma^2 n^2}{M^2}||w^*||^2_{\bH_{k^*:\infty}}  \right)  + \\
& + \frac{\sigma^2_{noise}}{100} \cdot  V_{k^*}(n,M)
%\left( \frac{k^*}{n}  + \frac{\gamma^2 n}{M^2} \sum_{j>k^*} \lam_j^2  \right) 
\;,
\end{align*}
where $ V_{k^*}(n,M)$ is defined in \eqref{eq:Vstar}.
\end{theorem}

The lower bound for the excess risk also decomposes into a bias part (first term)  and a part 
associated to the variance (second term). Comparing the bias with the upper bound for the bias from 
Theorem \ref{theo-main} shows that both are of the same order. Comparing the variances reveals that they are of the same 
order if 
\[  \frac{2\tau M^2\left(  ||w^*||^2_{\bI_{0:k^*}} +
 \gamma \frac{n}{M} || w^*||^2_{\bH_{k^*:\infty}} \right) }{\gamma  n (1-\gamma \tau \tr[\bH ]) } \lesssim 1 \;. \] 
In the next section, we will provide specific conditions and examples when this is satisfied.

%%%%%%%%%%%%%%%%%%%%%%%%%%%%%%%%%%%%%%%%%%%%%%%%%%%%%%%%%%%%%%%%%%%%%%%%%%%%%%%%%%
%%%%%%%%%%%%%%%%%%%%%%%%%%%%%%%%%%%%%%%%%%%%%%%%%%%%%%%%%%%%%%%%%%%%%%%%%%%%%%%%%%

\subsection{Fast Rates of convergence for specific distributions}
\label{subsec:rates}

We now consider two particular cases of data distributions, namely the \emph{spiked covariance model} (with local overparameterization) and the case where 
the eigenvalues of the second moment operator $\bH$ decay \emph{polynomially}. These are standard assumptions for the model, see e.g. 
\cite{ tsigler2020benign, zou2021benign, mucke2022data}. 
In both cases, we determine a range of the number of local nodes 
$M_n$ depending on the global sample size such that the bias is dominated by the variance. The final error is then of the order of the variance, if the number of local nodes grows sufficiently 
slowly with the sample size. The optimal\footnote{\emph{Optimal} in the sense of the maximal possible number of local nodes that balances bias and variance.} 
number exactly balances bias and variance.  

%\note{maybe talk about minimizing the rhs in $M$}

%\[ M_{opt} := \argmin_{M \in \mbn} \{  \bias (\w_M) + \var(\w_M)  \} \]

\vspace{0.2cm}

\begin{corollary}[Spiked Covariance Model]
\label{cor:spiked}
Suppose all assumptions of Theorem \ref{theo-main} are satisfied. 
Assume that $||w^*|| \leq R$ for some $R >0$ and $\bH \in \mbr^{d \times d}$. Let $d=\left( \frac{n}{M}\right)^q$ for some $q > 1$ and 
$\tilde d = \left( \frac{n}{M}\right)^r < d$ for some $0< r \leq 1$. Suppose the spectrum of $\bH$ satisfies 
\[ 
\lam_j = \begin{cases} \frac{1}{\tilde d}     &: \;j \leq \tilde d  \\  
                       \frac{1}{d - \tilde d} &:\; \tilde d + 1 \leq j \leq d \;.
          \end{cases}
 \]
If 
\[ M_n \leq \sqrt{\frac{\gamma(1- 2\gamma \tau  ) n }{ R^2}} \] 
then for any $n$ sufficiently large, we have 
\[ \mbe\left[ L(\w_{M_n}) \right] - L(w^*)  \; \leq \;  c \; \frac{1}{\gamma M_n}\; \left( \frac{M_n}{n}\right)^\nu \;, \]
where $\nu = \min\{ 1-r , q-1 \}$ and for some $c< \infty$, depending on $\tau, \gamma, \sigma$. 
\end{corollary}

\vspace{0.2cm}

Choosing the maximum number of local nodes $M_n \simeq \sqrt{n}$ gives the fast rate of order 
\[  \mbe\left[ L(\w_{M_n}) \right] - L(w^*)  \; \lesssim \;  \left( \frac{1}{n} \right)^{\frac{\nu+1}{2}} \;.   \]
for the excess risk.

\vspace{0.2cm}

\begin{corollary}[Polynomial Decay]
\label{cor:poly}
Suppose all assumptions of Theorem \ref{theo-main} are satisfied with 
$\gamma < \min \left\{ 1, \frac{1}{\tau \tr[\bH]} \right\}$. 
Assume that $||w^*|| \leq R$ for some $R >0$.  Suppose the spectrum\footnote{Note that the choice $\lam_j = j^{-(1+r)}$ ensures that $\tr[\bH] < \infty$.} 
of $\bH$ satisfies for some $r >0$ 
\[  \lam_j = j^{-(1+r)} \;. \] 
If 
\[  M_n \leq \left(\frac{\gamma }{R^2}\right)^{\frac{1+r}{2+r}} \cdot (\gamma n)^{\frac{1}{2+r}}  \;, \] 
then for any $n$ sufficiently large, we have 
\[ \mbe\left[ (\w_M) \right] - L(w^*)  \; \leq \;  c \; \frac{\gamma}{ M_n}\; \left( \frac{M_n}{n}\right)^{\frac{r}{1+r}} \;, \]
for some $c< \infty$, depending on $\tau, \gamma, \sigma$. 
\end{corollary}

\vspace{0.2cm}

Choosing the maximum number of local nodes $M_n \simeq n^{\frac{1}{2+r}}$ gives the fast rate of order 
\[  \mbe\left[ L(\w_{M_n}) \right] - L(w^*)  \; \lesssim \;  \left( \frac{1}{n} \right)^{\frac{r+1}{r+2}} \;.   \]
for the excess risk.

%\begin{remark}[General Source Condition]
%\note{remark more general source condition as in \cite{mucke2022data}, relaxation of the strong assumption $||w^*|| \leq R$}
%\end{remark}

%%%%%%%%%%%%%%%%%%%%%%% DISCUSSION %%%%%%%%%%%%%%%%%%%%%%%%%%%%%%%%%%%%%%%%%%

\subsection{Discussion}
\label{sec:main-discussion}

{\bf Comparison to single machine SGD.}  We compare the DSGD algorithm with the single machine SGD algorithm, i.e. when $M=1$. 
For this case, we recover the results from \cite{zou2021benign} under the same assumptions.  Our Corollaries \ref{cor:spiked}, \ref{cor:poly} show 
that the excess risk is dominated by the variance as long as $M$ grows sufficiently slowly with the sample size. But we can say even more: In the spiked covariance model, 
if $M_n \simeq n^\beta$ for  $\beta \in [0, 1/2]$, we see that DSGD performs as good as  single machine SGD, provided $\nu \leq 1$. Indeed, a direct 
comparison shows that
\begin{align*}
  \frac{1}{\gamma M_n}\; \left( \frac{M_n}{n}\right)^\nu 
&\simeq \frac{1}{\gamma n^\beta}\; \left( \frac{n^\beta}{n}\right)^\nu
\simeq \frac{1}{\gamma }\; \left( \frac{1}{n}\right)^\nu \;,
\end{align*} 
for any $\beta \in [0, 1/2]$ and $\nu \leq 1$. 
Recall that all our bounds are of optimal order, 
hence the relative efficiency remains of constant order until the critical threshold for $M_n$ 
is reached. 
\\
However, if $M_n$ is larger than the threshold, i.e. if $\beta \in (1/2, 1]$, then the bias term is dominating. In this case, the excess risk is of order 
\begin{align*}
\frac{2M^2}{ n^2}  ||w^*||^2_{\bH^\dagger_{0:k^*}}+ ||w^*||^2_{\bH_{k^*:\infty}}  \\
&\simeq \left( \frac{M_n}{n} \right)^{2-r} + \left( \frac{M_n}{n}\right)^q \\
&\simeq \left( \frac{n^\beta}{n} \right)^{2-r} + \left( \frac{n^\beta}{n}\right)^q\;,
\end{align*} 
being larger than the variance, see the proof of Corollary \ref{cor:spiked}, Appendix \ref{supp:proofsC}. 
\\
The same observations can be made for the setting in Corollary \ref{cor:poly} when the eigenvalues are polynomially decaying. 
If we let $M_n \simeq n^\beta$ with $\beta \in [0, 1/(2+r)]$, then the variance dominates and for all $r >0$, the test error satisfies 
\begin{align*}
  \frac{1}{\gamma M_n}\; \left( \frac{M_n}{n}\right)^{\frac{r}{r+1}} 
&\simeq \frac{1}{\gamma n^\beta}\; \left( \frac{n^\beta}{n}\right)^{\frac{r}{r+1}} 
\simeq \frac{1}{\gamma }\; \left( \frac{1}{n}\right)^{\frac{r}{r+1}}  \;.
\end{align*} 
We refer to Section\ref{sec:numerics} and Section \ref{sec:further-numerics} for some numerical experiments.

\[\]
{\bf Comparison to distributed learning in RKHSs.} We emphasize that 
all our results above hold for a  constant stepsize $0 < \gamma < \min \left\{ 1, \frac{1}{\tau \tr[\bH]} \right\}$. 
In particular, $\gamma$ does not depend on the number $M$ of local nodes. This result is line with 
the results for regularized distributed learning over reproducing kernel Hilbert spaces, 
see \citet{zhang2015divide, lin2017distributed, mucke2018parallelizing} 
and references therein.  
In this setting it is shown for a large class of spectral regularization methods\footnote{This class 
contains, among others,  gradient descent and accelerated methods like Heavy ball and Nesterov, 
ridge regression or PCA. } that the optimal regularization parameter $\lambda$ that leads to 
minimax optimal bounds, depends on the global sample size only and is of order 
$n^{-\alpha}$, $\alpha \in (0, 1]$. 
In particular, this parameter is chosen as in the single machine machine setting and each 
local subproblem is underregularized. 
This leads to a roughly constant  bias (unchanged by averaging) in the distributed setting, an increase in variance but 
averaging reduces the variance sufficiently to obtain optimal excess risk bounds. 
The same phenomenon occurs in our DSGD setting. On each local node the same stepsize $\gamma$ as for the $M=1$ case 
is applied.

\[\]
\noindent
{\bf Comparison to distributed ordinary least squares (DOLS).} 
We also compare our results with those recently obtained in \cite{mucke2022data} for DOLS in random design linear regression. 
The general observation in this work is that in the presence of overparameterization, the number of local nodes acts 
as a regularization parameter, balancing bias and variance. Recall that this is in contrast to what we observe for DSGD due to the 
implicit regularization.   
The optimal number of splits 
$M^{OLS}_{opt}$ depends on structural assumptions, i.e. eigenvalue decay and decay of 
the Fourier coefficients of $w^*$ (a.k.a. \emph{source condition}).  
\\
For the spiked covariance model,  
%in our Corollary \ref{cor:spiked}, 
the optimal number $M^{OLS}_{n}$ of DOLS is of order 
\[ M^{OLS}_{n} \simeq \left( \frac{d n^{3/2}}{d \cdot \tilde d} \right)^{2/5} \simeq n^{\frac{3-2r}{5-2r}} \;,  \]
see Corollary 3.14 in \cite{mucke2022data}. 
Comparing with our maximum number for $M_n \simeq n^{1/2}$ from our Corollary \ref{cor:spiked} we observe that $ M^{OLS}_{n} \lesssim M^{SGD}_n$ 
%\[ M^{OLS}_{n} \lesssim M^{SGD}_n \] 
if $\frac{1}{2} \leq r \leq 1$, i.e., DSGD allows for more parallelization in this regime.

For polynomially decaying eigenvalues $\lam_j \sim j^{1+r}$, $r >0$, the optimal number of data splits in Corollary 3.9  
in \cite{mucke2022data} scales as $ M^{OLS}_{n}  \simeq n^{1/3}$. 
%\[  M^{OLS}_{n}  \simeq n^{1/3} \;.\]
Compared to our result from Corollary \ref{cor:poly} we have 
\[  M^{SGD}_n \simeq n^{\frac{1}{2+r}} \lesssim n^{1/3} \]
for all $r\geq 1$. Thus, DOLS seems to allow more data splits under optimality guarantees for fast polynomial decay, 
i.e. large $r$.

%%%%%%%%%%%%%%%%%%%%%%%%%%%%%%%%%%%%%%%%%%%%%%%%%%%%%%%%%%%%%%%%%%%%%%%%%%%%%%%%%%%%%%%%%%%%%%%%%%%%%%%%%%%%
%%%%%%%%%   THE BENEFIT OF REGULARIZATION     %%%%%%%%%%%%%%%%%%%%%%%%%%%%%%%%%%%%%%%%%%%%%%%%%%%%%%%%%%%%%%%%%%%%%%%%%%%%%%%%%%%%%%
%%%%%%%%%%%%%%%%%%%%%%%%%%%%%%%%%%%%%%%%%%%%%%%%%%%%%%%%%%%%%%%%%%%%%%%%%%%%%%%%%%%%%%%%%%%%%%%%%%%%%%%%%%%%

\section{COMPARISON OF SAMPLE COMPLEXITY OF DSGD AND DRR}
\label{sec:benefit}

In this section we compare the distributed tail-averaged SGD estimator with the distributed Ridge Regression (RR) estimator
%ordinary least-squares (OLS) 
(see \cite{zhang2015divide, lin2017distributed, mucke2018parallelizing, sheng2020one} or \cite{tsigler2020benign} for RR in the single machine case). 
Recall that RR reduces to ordinary least-squares (OLS)
if the regularization parameter is set to zero. As a special case, we compare our results  to local OLS from 
\cite{mucke2022data} and analyze the benefit of implicit regularization of local SGD in the presence of local 
overparameterization.

We recall that for any $m \in [M]$, $\lam \geq 0$, the local RR estimates are defined by 
\[  \hat w_m^{\ols }(\lam ) = \bX_m^T( \bX_m \bX_m^T + \lam )^{-1} \bY_m \;.\]
The average is
\[ \olsest_n (\lam ) = \frac{1}{M} \sum_{m=1}^M \hat w_m^{\ols } \;. \]

We aim at showing that the excess risk of DSGD is upper bounded by the excess risk of DRR under suitable assumptions on the sample complexity. 
To this end, we first derive a lower bound for DRR to compare with. The proof follows by combining 
Proposition \ref{prop:low-bias-RR} and Proposition \ref{prop:low-var-RR} with Lemma \ref{prop:risk-local-RR}.   

\vspace{0.2cm}

\begin{assumption}
\label{ass:gauss}
The variable $\bH^{-1}x$ is sub-Gaussian and has independent components. 
\end{assumption}

\vspace{0.2cm}

Similarly to the bounds for DSGD, our bounds for DRR depend on the effective dimension 
\[ k^*_{\ols } := \min \left\{ k: \lam_{k+1} \leq \frac{ M\left( \lam + \sum_{j > k} \lam_j\right)}{bn} \right\}  \;, \]
for $\lam > 0$ and some $b >1$.

\vspace{0.2cm}

\begin{theorem}[Lower Bound Distributed RR]
\label{theo:lower-RR}
Suppose Assumption \ref{ass:gauss} holds and that 
$\bH$ is strictly positive definite with $\tr[\bH] < \infty$. Assume that $k^*_{\ols } \leq \frac{n}{c'M}$ 
for some $c'>1$.  
There exist constants $b,c >1$ such that the excess risk of the averaged RR estimator satisfies 
\begin{align*}
 \mbe\left[ L(\olsest_n (\lam ) ) \right] - L(w^*) &\geq ||w^*||^2_{\bH_{k^*_{\ols} : \infty}} 
+ \frac{M^2 \left( \lam + \sum_{j > k^*_{\ols}} \lam_j \right)^2}{c n^2} \cdot ||w^*||^2_{\bH^{-1}_{0:k^*_{\ols}}}   \\
& + \frac{\sigma^2}{c} \left( \frac{k^*_{\ols}}{n}  + \frac{n}{M^2} \cdot
\frac{ \sum_{j >k^*_{\ols} } \lam_j^2}{ ( \lam + \sum_{j >k^*_{\ols} } \lam_j)^2} \right) \;.
\end{align*}
\end{theorem}

\vspace{0.2cm}

We do our risk comparison particularly for tail-averaged DSGD and derive a bias-improved upper bound. The proof is given 
in Section \ref{sec:proof-upper-bound-tail-ave} and is an extension of Lemma 6.1 in \cite{zou2021benefits} to DSGD.

\vspace{0.2cm}

\begin{theorem}[Upper Bound Tail-averaged DSGD]
\label{theo:upp-tail-DSGD}
Suppose Assumption \ref{ass:well}  is satisfied. Let $\w_{M_{n}}$ denote the tail-averaged distributed estimator with $n$ training samples and assume $\gamma < 1/\tr[H]$. 
For arbitrary $k_1, k_2 \in [d]$ 
\begin{align*}
 \mbe\left[ L(\w_{M} ) \right] - L(w^*)  \; &=  \bias (\w_M) +   \var (\w_M) \; 
\end{align*}
with 
\begin{align*}
  \bias (\w_M) &\leq 
\frac{c_b M^2}{\gamma^2 n^2} \cdot \left| \left| \exp\left( -\frac{n}{M}\gamma \bH  \right) w^*\right| \right|^2_{\bH^{-1}_{0:k_1}}  
 +    || w^*||^2_{\bH_{k_1:\infty}}  \;, 
\end{align*} 
\[ \var (\w_M) \leq   c_v (1+R^2)\cdot \sigma^2   \left( \frac{k_2 }{n} + \frac{n \gamma^2}{M^2} \cdot \sum_{j > k_2}\lam_j^2  \right)     \;, \]
for some universal constants $c_b , c_v>0$.
\end{theorem}

\vspace{0.2cm}

To derive the risk comparison we fix a sample size $n_\ols$ and $n_{\sgd}$ for DRR and tail-averaged DSGD, resp., 
and derive conditions on the sample complexities such that individually, the bias and variance of DSGD is upper bounded by the bias and 
variance of DRR, respectively. Combining then both of the above theorems finally leads to the risk comparison result. 
%A detailed computation is given in Section \ref{sec:comparison}.

\vspace{0.2cm}

\begin{theorem}[Comparison DSGD with DRR]
\label{theo:comparison}
Let $\w_{M_{n_{\sgd}}}$ denote the tail-averaged distributed estimator with $n_{\sgd}$ training samples. 
Let further $\olsest_{n_\ols} (\lam )$ denote the 
distributed RR estimator with $n_{\ols}$ training samples and with regularization parameter $\lam\geq 0$. 
Suppose all assumptions from Theorems \ref{theo:lower-RR} ,\ref{theo:upp-tail-DSGD} are satisfied.  
%\ref{ass:well} is satisfied \note{also gaussian distribution}. 
There exist constants $b, c >1$ and $0 < L_{\lam, \gamma} \leq L'_{\lam, \gamma} $ such that for  
$C^*:=c\left(1+\frac{||w^*||^2}{\sigma^2}\right)$, 
\[  C_\lam^* := \lam + \sum_{j >k^*_{\ols}} \lam_j \;, \]
\begin{equation}
\label{eq:stepsize}
\gamma < \min\left\{ \frac{1}{\tr[H]} , \frac{1}{\sqrt{c}C^* C_\lam^*} \right\}
\end{equation} 
and 
\begin{equation*}
 L_{\lam, \gamma}\cdot  n_{\ols}     \leq   n_{\sgd}    \leq   L'_{\lam, \gamma}\cdot    n_{\ols}
\end{equation*} 
the excess risks of DSGD and DRR satisfy 
\begin{align}
\label{eq:risk-comp}
 \mbe\left[ L(\w_{M} ) \right] - L(w^*)  \; &\leq \;  \mbe\left[ L(\olsest_{n_\ols} (\lam ) ) \right] - L(w^*)   \;.
\end{align}
The constants $L_{\lam, \gamma} , L'_{\lam, \gamma} $ are explicitly given by 
\[L_{\lam, \gamma} = \max\left\{  C^* , \frac{\sqrt{c(1-\gamma \lam_{k^*_{\ols}})}}{\gamma C_\lam^*}  \right\} \;, \]
\[  L'_{\lam, \gamma}  = \frac{1}{C^* \gamma^2 (C_\lam^*)^2}   \;.\]
\end{theorem}

Note that in the above Theorem, assumption \eqref{eq:stepsize} on the stepsize ensures 
that $0 < L_{\lam, \gamma} \leq L'_{\lam, \gamma}$. 
\\ 
We next show that under an appropriate condition on the amount regularization, the sample complexities are indeed of the same order.

To ensure that $ n_{\ols} \lesssim n_\sgd$ we need to require that 
\[  1 \lesssim   L_{\lam_{n_\ols}, \gamma} = \max\left\{  C^* , \frac{\sqrt{c(1-\gamma \lam_{k^*_{\ols}})}}{\gamma C_{\lam_{n_\ols}}^*}  \right\}  \;.  \]
%with $ C_\lam^* := \lam + \sum_{j >k^*_{\ols}} \lam_j $. 
Recall that 
\[ \gamma < \min\left\{ \frac{1}{\tr[H]} , \frac{1}{\sqrt{c}C^* C_{\lam_{n_\ols}}^*} \right\}\;, \]
and that $1-\gamma \lam_{k^*_{\ols}} < 1$. A short calculation shows that 
\[ 1 \lesssim  \frac{\sqrt{c(1-\gamma \lam_{k^*_{\ols}})}}{\gamma C_{\lam_{n_\ols}}^*}   \]
if 
\[ \gamma \left( \lam_{n_\ols} + \sum_{j >k^*_{\ols}} \lam_j \right) \lesssim 1 \;. \]

Furthermore, to ensure that  $ n_\sgd \lesssim n_{\ols}$ we have to require that 
\[   L'_{\lam_{n_\ols} , \gamma}  = \frac{1}{C^* \gamma^2 (C_{\lam_{n_\ols}}^*)^2}  \lesssim 1 \;. \] 
This is satisfied if 
\[ 1 \lesssim \gamma \cdot C_{\lam_{n_\ols}}^* = \gamma  \left( \lam_{n_\ols} + \sum_{j >k^*_{\ols}} \lam_j \right) \;.  \]
We summarize our findings in the following:

\begin{corollary}
\label{cor:same-order}
Suppose all assumptions of Theorem \ref{theo:comparison} are satisfied. If 
\begin{equation}
\label{eq:comp-cond-1}
\gamma \left( \lam_{n_\ols} + \sum_{j >k^*_{\ols}} \lam_j \right) \simeq  1
\end{equation}
%and 
%\begin{equation}
%\label{eq:comp-cond-2}
%1 \lesssim \gamma \cdot C_{\lam_{n_\ols}}^*  
%\end{equation}
holds, then the sample complexities of DSGD and DRR are of the same order, i.e. 
\[  n_\sgd \simeq n_\ols  \]
and 
\[  \mbe\left[ L(\w_{M_{n_\sgd}} ) \right] - L(w^*)  \; \leq \;  \mbe\left[ L(\olsest_{n_\ols} (\lam_{n_\ols} ) ) \right] - L(w^*)  \;. \] 
\end{corollary}

\vspace{0.2cm}

%{\bf Example:}
\begin{example}[Spiked Covariance Model]
We show that condition \eqref{eq:comp-cond-1} is satisfied in the spiked covariance model from Corollary \ref{cor:spiked} under a suitable choice for 
$\lam_{n_\ols}$ and $M_{n_\sgd}$. 
%Recall that the maximum number of local nodes scales as 
%\[ M_{n_\sgd} \simeq \sqrt{\gamma n_\sgd} \;. \]
Here, we assume that with 
\[ M_{n_\sgd} = M_{n_\ols} \simeq n_\ols^{\frac{3-2r}{5-2r}} \;, \]
for $1/2 \leq r \leq 1$, see our discussion in Section \ref{sec:main-discussion} (comparison with DOLS).
A short calculation shows that 
\[  k^*_{\ols}  \simeq \tilde d \simeq \left( \frac{n_\ols}{M_{n_\ols}}\right)^r \simeq n^{\frac{2r}{5-2r}}_\ols \;. \]
Moreover, for $\lam_{n_\ols} \simeq n_{\ols}^{-\zeta }$, $\zeta \geq 0$ and $\gamma = const.$ , we have 
\[ \gamma \left( \lam_{n_\ols} + \sum_{j >k^*_{\ols}} \lam_j \right) 
 \simeq  \gamma \left( n_{\ols}^{-\zeta } + 1 \right) \simeq 1 \;.\]
Hence,  for a wide range of regularization, the condition \eqref{eq:comp-cond-1}  is met and the SCs of DSGD and DRR 
in the spiked covariance model are of the same order. 
\end{example}

Our result shows that DSGD performs better than DRR/ DOLS if the sample complexity (SC) of SGD differs from the SC of RR/OLS by no more than a 
constant. This constant depends on the amount of regularization $\lambda$, the stepsize $\gamma$ and the tail behavior of the eigenvalues of the Hessian.

Our bound slightly differs from  \cite{zou2021benefits} for the case $M=1$ in  two respects: We scale our SC such that the constant in \eqref{eq:risk-comp} is equal to one 
while  \cite{zou2021benefits} show that both risks are of the same order (with a constant larger than one). Second, we also show that the SC of DSGD is upper bounded 
by a factor of the SC of DRR/DOLS while  \cite{zou2021benefits} only derive a lower bound. 
%However, we remark that $n_{\sgd}$ in our Theorem is larger that $n_{\ols}$ as the constant $L_{\lam, \gamma} \geq 1$. A look onto Figure \ref{fig:1}  
%reveals that optimally tuned  DSGD may perform better than optimally tuned DRR even with the same or smaller sample size for certain problem instances. This 
%suggests that our bound may be refined. 

%%%%%%%%%%%%%%%%%%%%%%%%%%%%%%%%%%%%%%%%%%%%%%%%%%%%%%%%%%%%%%%%%%%%%%%%%%%%%%%%%%%%%%%%%%%%%%%%%%%%%%%%%%%%
%%%%%%%%%   Numerical Experiments     %%%%%%%%%%%%%%%%%%%%%%%%%%%%%%%%%%%%%%%%%%%%%%%%%%%%%%%%%%%%%%%%%%%%%%%%%%%%%%%%%%%%%%
%%%%%%%%%%%%%%%%%%%%%%%%%%%%%%%%%%%%%%%%%%%%%%%%%%%%%%%%%%%%%%%%%%%%%%%%%%%%%%%%%%%%%%%%%%%%%%%%%%%%%%%%%%%%

\section{NUMERICAL EXPERIMENTS}
\label{sec:numerics}

We illustrate our theoretical findings with experiments on simulated and real data. The reader may find additional experiments in Section \ref{sec:further-numerics}.

{\bf Simulated Data.}
%\subsection{Simulated Data}
%\label{sec:num-simu}

\begin{figure}[h]
\centering
\includegraphics[width=0.49\columnwidth, height=0.3\textheight]{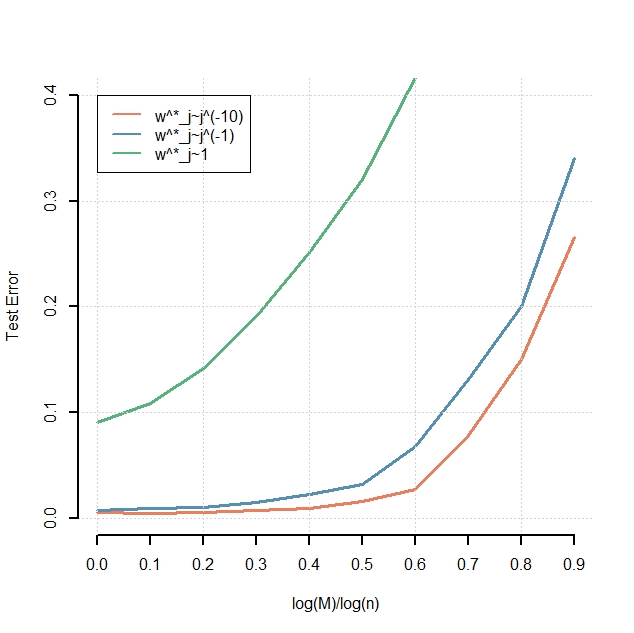}
\includegraphics[width=0.49\columnwidth, height=0.3\textheight]{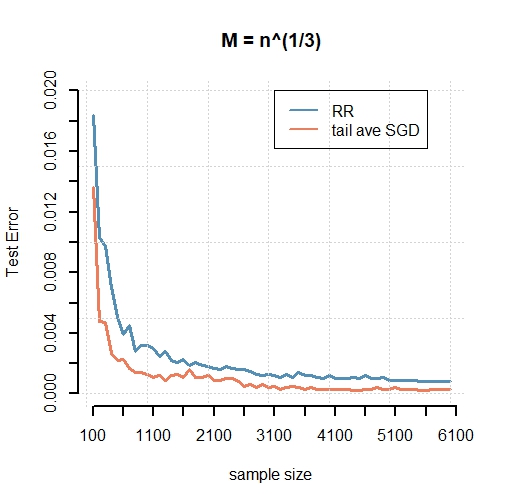}
\caption{{\bf Left:} Test error for DSGD with $\lam_j = j^{-2}$ for different sources $w^*$ as a function of $M$. 
{\bf Right:} Comparison of optimally tuned tail-ave DSGD with DRR with $\lam_j = j^{-2}$, $w^*_j=j^{-10}$, $M_n=n^{1/3}$.}
\label{fig:1}
\end{figure}

\begin{figure}[h]
\centering
\includegraphics[width=0.49\columnwidth, height=0.3\textheight]{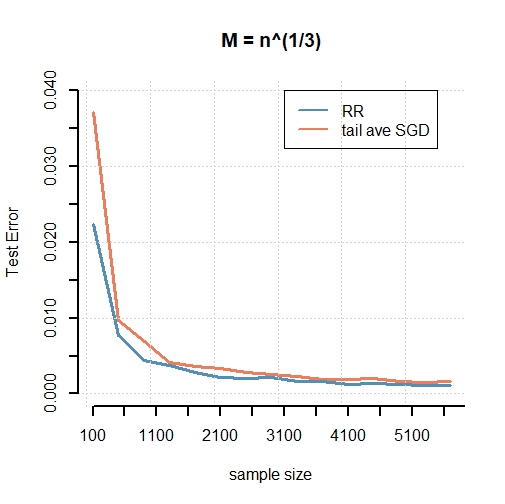}
\includegraphics[width=0.49\columnwidth, height=0.3\textheight]{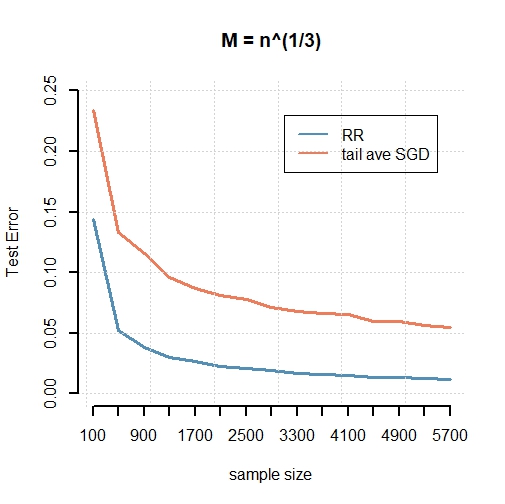}
\caption{Comparison of optimally tuned tail-ave DSGD with DRR with $\lam_j = j^{-2}$ for different sources $w^*$, with $\lam_j = j^{-2}$ and $M_n=n^{1/3}$. 
{\bf Left:} $w^*_j=j^{-1}$
{\bf Right:} $w^*_j=1$.}
\label{fig:4}
\end{figure}

\noindent
In a first experiment in Figure \ref{fig:1} (left) we analyze the test error of DSGD as a function of the local nodes $M$. We generate $n=500$ i.i.d. 
training data with $x_j \sim \cN(0, \bH)$ with 
mildly overparameterization $d=700$. The target $w^*$ satisfies three different decay conditions $w^*_j = j^{-\alpha}$, $\alpha \in \{0,1,10\}$. 
The eigenvalues of $\bH$ follow a polynomial decay $\lam_j = j^{-2}$. The local nodes  satisfy $M_n = n^{\beta}$, $\beta \in \{0, 0.1, ..., 0.9\}$. According to 
Corollary \ref{cor:poly} we see that a fast decay of $w^*_j$ (i.e. a smaller norm $||w^*||$) allows for more parallelization until the test error blows up. 
\\
In a second experiment we compare the sample complexity of optimally tuned tail-averaged DSGD and DRR for different sources $w^*$, 
see Figures \ref{fig:1} (right), \ref{fig:4}. Here, the data are generated as 
above with $d=200$, $\lam_j = j^{-2}$ and $w^*_j=j^{-\alpha}$, $\alpha \in \{0, 1, 10\}$. 
The number of local nodes is fixed at $M_n=n^{1/3}$ for each $n \in \{100, ..., 6000\}$. 
For this problem instance, DSGD may perform even better than DRR for sparse targets ($\alpha=10$), 
i.e.,  DSGD achieves the same accuracy as DRR with less samples in this regime. For less sparse targets $\alpha=1$, the sample complexities of DSGD and DRR 
are comparable while for non-sparse targets ($\alpha=0$), DRR outperforms DSGD.

%%%%%%%%%%%%%%%%%%%%%%%%%%%%%%%%%%%%%%%%%%%%%%%%%%%%%%%%%%%%%%%%%%%%%%%%%%%%%%%
%%%%%%%%% REAL DATA 
%%%%%%%%%%%%%%%%%%%%%%%%%%%%%%%%%%%%%%%%%%%%%%%%%%%%%%%%%%%%%%%%%%%%%%%%%%%%%%%

%\subsection{Real Data}
{\bf Real Data.}
To analyze the performance of DSGD on real data, we considered the classification problem of the Gisette data set\footnote{http://archive.ics.uci.edu/ml/datasets/Gisette}, containing pictures of the digits four and nine. We used the first $3000$ samples of the original train data set for training and the second $3000$ samples for evaluation. 
The feature dimension of one picture is  $d=5000$. Hyper-parameters had been fine-tuned on the validation data set to achieve the best performance. 
\begin{figure}[h!]
\centering
\includegraphics[width=0.45\columnwidth, height=0.22\textheight]{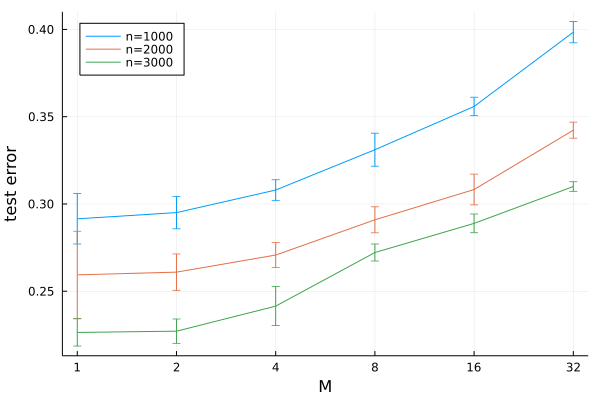}
\includegraphics[width=0.45\columnwidth, height=0.22\textheight]{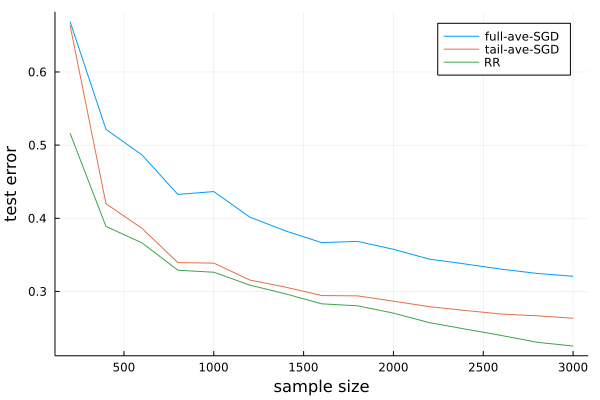}
\caption{{\bf Left:} Test error for DSGD with $n = 1000, 2000, 3000$ and different $M$. 
{\bf Right:} Comparison of DSGD with DRR for $M_n=n^{1/4}$.}
\label{fig:2}
\end{figure}
The first experiment in Figure \ref{fig:2}(left) again analyzes the test error of DSGD as a function of the local nodes $M$. 
Because the feature dimension is quite large, the optimal stepsize is small ($\gamma\sim 10^{-10}$). Theorem \ref{theo:lower} therefore explains why  in our example the bias-term and thus the test error grows rather quickly with the number of local nodes. In Figure \ref{fig:2}(right) we compare DRR with tail- and full-averaged DSGD. We observe that DRR slightly outperforms DSG. According to Theorem \ref{theo:comparison}, we need sparsity for $w^*$ so that DSGD can keep up with DRR. This might be not the case for the Gisette data set.

%%%%%%%%%%%%%%%%%%%%%%%%%%%%%%%%%%%%%%%%%%%%%%%%%%%%%%%%%%%%%%%%%%%%%%%%%%%%%%%%%%%%%%%%%%%%%%%%%%%%%%%%%%%%
%%%%%%%%%   CONCLUSION    %%%%%%%%%%%%%%%%%%%%%%%%%%%%%%%%%%%%%%%%%%%%%%%%%%%%%%%%%%%%%%%%%%%%%%%%%%%%%%%%%%%%%%
%%%%%%%%%%%%%%%%%%%%%%%%%%%%%%%%%%%%%%%%%%%%%%%%%%%%%%%%%%%%%%%%%%%%%%%%%%%%%%%%%%%%%%%%%%%%%%%%%%%%%%%%%%%%

\section{Summary}

We analyzed the performance of distributed constant stepsize (tail-) averaged SGD for linear regression in an overparameterized regime. 
We find that the relative efficiency as a function of the number of workers  remains largely unchanged until a certain threshold is reached. This threshold depends 
on the structural assumptions imposed by the problem at hand (eigenvalue decay of the Hessian $\bH$ and the norm of the target $w^*$). This is in contrast to 
distributed OLS without any implicit or explicit regularization with local overparameterization, where the number of workers itself acts as a regularization parameter, 
see Figure \ref{fig:3} in Appendix \ref{sec:further-numerics}. 
\\
We also compared the sample complexity of DSGD and DRR and find that the sample complexity of DSGD remains within constant factors of the sample complexity of DRR. 
For some problem instances, tail-averaged SGD may outperform DRR, i.e., achieves the same or better accuracy with less samples. Our bound is not sharp and may be 
improved in future research.

%%%%%%%%%%%%%%%%%%%%%%%%%%%%%%%%%%%%%%%%%%%%%%%%%%%%%%%%%%%%%%%%%%%%%%%%%%%%%%%%%%%%%%%%%%%%%%%%%%%
%\vspace{2cm}
%\noindent
%{\Large \bf Acknowledgments}
%\\
%\\
%TEST

%\bibliographystyle{plain}
%\bibliographystyle{abbrv}
\bibliographystyle{alpha}
\bibliography{bib_distributed}

%%%%%%%%%%%%%%%%%%%%%%%%%%%%%%%%%%%%%%%%%%%%%%%%%%%%%%%%%%%%%%%%%%%%%%%%%%%%%%%%%%%%%%%%%%%%%%%%%%%%%%%%%%%%
%%%%%%%%%    SUPPLEMENT    %%%%%%%%%%%%%%%%%%%%%%%%%%%%%%%%%%%%%%%%%%%%%%%%%%%%%%%%%%%%%%%%%%%%%%%%%%%%%%%%%%%%%%
%%%%%%%%%%%%%%%%%%%%%%%%%%%%%%%%%%%%%%%%%%%%%%%%%%%%%%%%%%%%%%%%%%%%%%%%%%%%%%%%%%%%%%%%%%%%%%%%%%%%%%%%%%%%

\newpage 
\appendix

\paragraph{Notation.} 
By $\cL(\cH_1, \cH_2)$ we denote the space of bounded linear operators between real Hilbert spaces $\cH_1$, $\cH_2$ with operator norm $||\cdot ||$. 
We write $\cL(\cH, \cH) = \cL(\cH)$. For $\bA \in \cL(\cH)$ we denote by $\bA^T$ the adjoint operator. 
For two PSD operators on $\cH$ we write $\bA \preceq \bB$ if $\inner{(\bA - \bB) v , v} \geq 0$ for all $v \in \cH$. 
We further let $\inner{\bA , \bB}_{op} = \tr[\bA ^T \bB]$.

%%%%%%%%%%%%%%%%%%%%%%%%%%%%%%%%%%%%%%%%%%%%%%%%%%%%%%%%%%%%%%%%%%%%%%%%%%%%%%%%%%%%%%%%%%%%%%%%
%%%%%%%%%%%%%%%%%%%%% Proofs Local SGD
%%%%%%%%%%%%%%%%%%%%%%%%%%%%%%%%%%%%%%%%%%%%%%%%%%%%%%%%%%%%%%%%%%%%%%%%%%%%%%%%%%%%%%%%%%%%%%%%

\section{PROOFS SECTION 3 (BOUNDS FOR DSGD)}

%%%%%%%%%%%%%%%%%%%%%%%%%%%%%%%%%%%%%%%%%%%%%%%%%%%%%%%%%%%%%%%%%%%%%%%%%%%%%%%%%%%%%%%%%%%%%%%%
%%%%%%%%%%%%%%%%%%%%% Proofs Upper Bound 
%%%%%%%%%%%%%%%%%%%%%%%%%%%%%%%%%%%%%%%%%%%%%%%%%%%%%%%%%%%%%%%%%%%%%%%%%%%%%%%%%%%%%%%%%%%%%%%%

\subsection{Proofs Upper Bound}
\label{supp:proofsA}

%%%%%%%%%%%%%%%%%%%%%%%%%%%%%%%%%%%%%%%%%%%%%%%%%%%%%%%%%%%%%%%%%%%%%%%%
%%% Bias-Variance Decomposition 
%%%%%%%%%%%%%%%%%%%%%%%%%%%%%%%%%%%%%%%%%%%%%%%%%%%%%%%%%%%%%%%%%%%%%%%%

\subsubsection{Bias-Variance Decomposition}
We will use an iterative bias-variance-decomposition which has been
extensively studied before in the non distributed case (see \cite{Kakade16}, \cite{zou2021benign}). First we need a couple of definitions.

\begin{itemize}
\item[-)]\textbf{Centered local iterates:} Set $\eta^{(m)}_t := w^{(m)}_t - w^*$ and  \\
\[  \bar \eta^{(m)}_n := \frac{M}{n}\sum_{t=1}^{n/M}  \eta_t^{(m)} \;, \quad \bar{\bar{\eta}}_M:= \frac{1}{M} \sum_{m=1}^M  \bar \eta^{(m)}_n .\]\vspace{0.1cm}
\item[-)]\textbf{Local bias:} For $m=1,...,M$ we set $b_1^{(m)} = w_1-w^*$, \\\\
\[ b_t^{(m)}:= (\bI - \gamma x_t^{(m)} \otimes x_t^{(m)})b^{(m)}_{t-1}\;, \quad t=2,...,\frac{n}{M} \; \]
\vspace{0.1cm} 
\[  \bar b^{(m)}_n := \frac{M}{n}\sum_{t=1}^{n/M}  b_t^{(m)} \;, \quad \bb_M:= \frac{1}{M} \sum_{m=1}^M  \bar b^{(m)}_n .\]
\vspace{0.1cm}
\item[-)]\textbf{Local variance:} For $m=1,...,M$ we set $v_1^{(m)} = 0$ and 
 \vspace{0.2cm}
\[ v_t^{(m)}:= (\bI - \gamma x_t^{(m)}\otimes x_t^{(m)})v^{(m)}_{t-1} + \gamma \epsilon_t^{(m)} x_t^{(m)} \;, \quad t=2,...,\frac{n}{M} \;, \]
\vspace{0.1cm}
\[  \bar v^{(m)}_n := \frac{M}{n}\sum_{t=1}^{n/M}  v_t^{(m)} \;, \quad \vv_M:= \frac{1}{M} \sum_{m=1}^M  \bar v^{(m)}_n, \]
\vspace{0.2cm}
where we let $\epsilon_t^{(m)} := y_t^{(m)} - \inner{x_t^{(m)} , w^*}$.
\end{itemize}

\vspace{0.2cm}

Note that for any $m=1,...,M$ and $t \geq 1$ one has 
\begin{equation}
\label{eq:expec-bias}
\mbe[ b_{t+1}^{(m)}] = \mbe[ \mbe[  b_{t+1}^{(m)} |  b_{t}^{(m)} ] ] = \mbe[ \mbe[ (\bI - \gamma x^{(m)}_{t+1} \otimes x^{(m)}_{t+1})b^{(m)}_{t} |  b_{t}^{(m)} ] ] 
= (\bI - \gamma \bH) \mbe[ b_{t}^{(m)}] \;.
\end{equation}

Moreover, from B.4 in \cite{zou2021benign}, we find 
\begin{equation}
\label{eq:varit}
\mbe[ v_{t+1}^{(m)} ] = (\bI - \gamma H)\;  \mbe[v_{t}^{(m)}]  = (\bI - \gamma H)^t \; \mbe [v^{(m)}_1 ] = 0\;.
\end{equation}

\vspace{0.2cm}

It is easy to see that $\eta^{(m)}_t = b^{(m)}_t + v^{(m)}_t$ and therefore
\vspace{0.2cm}
\begin{align}
\bar{\bar{\eta}}_M = \bb_M+\vv_M .\label{zerlegung}
\end{align}

\vspace{0.2cm}

\begin{lemma}	
\label{lem:bias-variance-deco}
Define 
\[\bias (\w_M) := \frac{1}{2}\inner{ \bH , \mbe\left[ \bb_M \otimes \bb_M \right] }_{op} \;, \quad 
\var(\w_M) :=   \frac{1}{2}\inner{ \bH , \mbe\left[ \vv_M \otimes \vv_M \right] }_{op} \;. \]
\begin{itemize}
\item[$a)$] We have the following decomposition for the excess risk,
\[ \mbe\left[ L(\w_M) \right] - L(w^*)  \leq \left(\sqrt{\bias (\w_M)}  + 
\sqrt{\var(\w_M)}\right)^2.\] 
\item[$b)$] Suppose the model noise $\epsilon_t^{(m)}$ is well-specified, i.e., 
$\epsilon_t^{(m)} := y_t^{(m)} - \inner{x_t^{(m)} , w^*}$ and $x_t^{(m)}$ are independent and $\mathbb{E}[\epsilon_t^{(m)}]=0$, then we have the following equality for the excess risk,
\begin{align*}
\mbe\left[ L(\w_M) \right] - L(w^*)  
&= \bias (\w_M)  + \var(\w_M) \;.
\end{align*}
\end{itemize}
\end{lemma}

\vspace{0.2cm}

\begin{proof}[Proof of Lemma \ref{lem:bias-variance-deco}]
The proof strategy is similar to the non distributed case (see \cite{zou2021benign}, Lemma B2 and Lemma C1). For completeness we included it here.\\
$a)$ By definition of the excess risk we have
\begin{align*}
\label{eq:exrisk_norm}
L(\w_M)-L(w^*)
&=\frac{1}{2}\int_\mathcal{H}\langle\w_M-w^*,x\rangle^2\;\mathbb{P}_\mathbf{x}(d\mathbf{x})\notag\\
&=\frac{1}{2}\langle \bH(\w_M-w^*),\w_M-w^*\rangle\notag\\
&=\frac{1}{2}\|\mathbf{H}^{\frac{1}{2}}(\w_M-w^*)\|^2\\
&=\frac{1}{2}\left\|\bb+\vv\right\|_\bH^2,
\end{align*}
where we used (\ref{zerlegung}) for the last equality.
Using Cauchy-Schwarz inequality we obtain
\begin{align*}
\mathbb{E}[L(\w_M)-L(w^*)]
&\leq\left(\sqrt{\frac{1}{2}\mathbb{E}\left\|\bb\right\|_\bH^2}+\sqrt{\frac{1}{2}\mathbb{E}\left\|\vv\right\|_\bH^2}\right)^2\\
&=\left(\sqrt{\frac{1}{2}\inner{ \bH , \mathbb{E}\left[\bb_M \otimes \bb_M \right] }_{op} }+ 
\sqrt{\frac{1}{2}\inner{ \bH , \mathbb{E}\left[\vv_M \otimes \vv_M \right] }_{op}}\right)^2
\end{align*}

$b)$ Set $P_t^{(m)}=\bI - \gamma x_t^{(m)} \otimes x_t^{(m)} $. Note that we have
\begin{align*}
b_t^{(m)}=\prod_{k=1}^t P_k^{(m)} b_0^{(m)},    && v_t^{(m)}=\gamma \sum_{i=1}^t\prod_{j=i+1}^t \epsilon_i^{(m)}P_j^{(m)} x_i^{(m)} .
\end{align*}
By assumption, we therefore have for all $s,t \leq n/M$ and $ m,m'\leq M$,
\begin{align*}
\mbe\left[ b_s^{(m)}\otimes v_t^{(m')} \right] &= \gamma \mbe\left[ \prod_{k=1}^s P_k^{(m)} b_0^{(m)}\otimes \sum_{i=1}^t\prod_{j=i+1}^t \epsilon_i^{(m')}P_j^{(m')} x_i^{(m')} \right]\\
&= \gamma \sum_{i=1}^t\mbe\left[ \prod_{k=1}^s P_k^{(m)} b_0^{(m)}\otimes \prod_{j=i+1}^t P_j^{(m')} x_i^{(m')} \right] \mathbb{E}[\epsilon_i^{(m')}]=0.
\end{align*}
This implies 
\begin{align}
\mbe\left[ \bb_M\otimes \vv_M \right] = 0. 
\end{align}
From (\ref{zerlegung}) we therefore have
\begin{align}
\mbe\left[ \bar{\bar{\eta}}_M\otimes \bar{\bar{\eta}}_M \right]= \mbe\left[ \bb_M\otimes \bb_M\right]+\mbe\left[ \vv_M\otimes \vv_M \right]\label{biasvarterm2}
\end{align}

Finally, by definition of the excess risk we have
\begin{align}
%\label{eq:exrisk_norm}
\mathbb{E}[L(\w_M)-L(w^*)]
&=\frac{1}{2}\mathbb{E}\left[\int_\mathcal{H}\langle\w_M-w^*,x\rangle^2\;\mathbb{P}_\mathbf{x}(d\mathbf{x})\right]\notag\\
&=\frac{1}{2}\mathbb{E}\left[\langle \bH(\w_M-w^*),\w_M-w^*\rangle\right]\notag\\
&=\frac{1}{2}\inner{ \bH , \mbe\left[ \bar{\bar{\eta}}_M\otimes \bar{\bar{\eta}}_M \right] }_{op}\\
&=\bias (\w_M) + \var(\w_M) ,
\end{align}
where we used (\ref{biasvarterm2}) for the last equality.

\end{proof}

%In the following we name 

%\begin{align*}
%\text{Bias}(\w_M):= \frac{1}{2}\inner{ \bH , \mathbb{E}\left[\bb_M \otimes \bb_M \right] }_{op},&&\text{
%Variance}(\w_M):= \frac{1}{2}\inner{ \bH , \mathbb{E}\left[\vv_M \otimes \vv_M \right] }_{op}.
%\end{align*}

%%%%%%%%%%%%%%%%%%%%%%%%%%%%%%%%%%%%%%%%%%%%%%%%%%%%%%%%%%%%%%%%%%%%%%%%
%%% Upper Bound Bias and Variance
%%%%%%%%%%%%%%%%%%%%%%%%%%%%%%%%%%%%%%%%%%%%%%%%%%%%%%%%%%%%%%%%%%%%%%%%

\subsubsection{Upper Bound}

For the non distributed case \cite{zou2021benign} (see Lemma B.11 and Lemma B.6 ) already established upper bounds. 
More precisely we have for the local bias and variance term: \vspace{0.2cm}

\begin{proposition}
\label{prop-upperbound}
Set  $k^*=\max \left\{k: \lambda_k \geq \frac{M}{n \gamma}\right\}$. If the step size satisfies $\gamma<1 /(\tau \operatorname{tr}(\mathbf{H}))$, we have for every $m\in[M]$:
\begin{itemize}
\item[a)]  Under Assumption \ref{ass:second-moment} and \ref{ass:fourth-moment}, it holds that 
\begin{align*}
 \text{Bias}\left(\bar w^{(m)}_{\frac{n}{M}}\right) :=&\frac{1}{2}\inner{ \bH , \mathbb{E}\left[b_t^{(m)} \otimes b_t^{(m)}\right] }_{op}\\[3pt]
			\leq & \frac{M^2}{\gamma^2 n^2} \cdot
            \left\|\mathbf{w}_0-\mathbf{w}^*\right\|_{\mathbf{H}_{0: k^*}^{-1}}^2+\left\|\mathbf{w}_0-\mathbf{w}^*\right\|_{\mathbf{H}_{k^*: \infty}^2} \\
&+\frac{2 \tau M^2\left(\left\|\mathbf{w}_0-\mathbf{w}^*\right\|_{\mathbf{I}_{0: k^*}}^2+\frac{n}{M} \gamma\left\|\mathbf{w}_0-\mathbf{w}^*\right\|_{\mathbf{H}_{k^*: \infty}}^2\right)}{\gamma n(1-\gamma \tau \operatorname{tr}(\mathbf{H}))} \cdot\left(\frac{k^*}{n}+\frac{n}{M^2} \gamma^2 \sum_{i>k^*} \lambda_i^2\right).
\end{align*}

\item[b)] Under Assumptions \ref{ass:second-moment} - \ref{ass:noise}, it holds that

\begin{align*}
 \text{Var}\left(\bar w^{(m)}_{\frac{n}{M}}\right):=\frac{1}{2}\inner{ \bH , \mathbb{E}\left[\bar v^{(m)}_n \otimes \bar v^{(m)}_n \right] }_{op}\leq \frac{\sigma^2}{1-\gamma \tau \operatorname{tr}(\mathbf{H})}\left(\frac{k^*M}{n}+\gamma^2 \frac{n}{M} \cdot \sum_{i>k^*} \lambda_i^2\right).
\end{align*}
\end{itemize}
\end{proposition}

\vspace{0.2cm}

\begin{lemma}
\label{distupbound}
Set  $k^*=\max \left\{k: \lambda_k \geq \frac{M}{n \gamma}\right\}$. If the step size satisfies $\gamma<1 /(\tau \operatorname{tr}(\mathbf{H}))$, we have for every $m\in[M]$:
\begin{itemize}
\item[a)]   Under Assumption \ref{ass:second-moment} and \ref{ass:fourth-moment}, it holds that 
\begin{align*}
\text{Bias}(\w_M) \leq & \frac{M^2}{\gamma^2 n^2} \cdot
            \left\|\mathbf{w}_0-\mathbf{w}^*\right\|_{\mathbf{H}_{0: k^*}^{-1}}^2+\left\|\mathbf{w}_0-\mathbf{w}^*\right\|_{\mathbf{H}_{k^*: \infty}^2} \\
&+\frac{2 \tau M^2\left(\left\|\mathbf{w}_0-\mathbf{w}^*\right\|_{\mathbf{I}_{0: k^*}}^2+\frac{n}{M} \gamma\left\|\mathbf{w}_0-\mathbf{w}^*\right\|_{\mathbf{H}_{k^*: \infty}}^2\right)}{\gamma n(1-\gamma \tau \operatorname{tr}(\mathbf{H}))} \cdot\left(\frac{k^*}{n}+\frac{n}{M^2} \gamma^2 \sum_{i>k^*} \lambda_i^2\right).
\end{align*}

\item[b)] Under Assumptions \ref{ass:second-moment} - \ref{ass:noise} , it holds that

\begin{align*}
\text{Var}(\w_M)\leq \frac{\sigma^2}{1-\gamma \tau \operatorname{tr}(\mathbf{H})}\left(\frac{k^*}{n}+\gamma^2 \frac{n}{M^2} \cdot \sum_{i>k^*} \lambda_i^2\right).
\end{align*}
\end{itemize}
\end{lemma}

\vspace{0.2cm}

\begin{proof}[Proof of Lemma \ref{distupbound} ]
$a)$ For the Bias-term we simply use 

	\begin{align}\label{eq:bias-upp-uniform}
	\text{Bias}(\w_M)=\frac{1}{2}\mathbb{E}\left\|\bb\right\|_\bH^2=\frac{1}{2}\mathbb{E}\left\|\frac{1}{M}\sum_{m=1}^M \bar{b_M}_n^{(m)}\right\|_\bH^2\leq \frac{1}{M}\sum_{m=1}^M \frac{1}{2}\mathbb{E}\left\| \bar{b}_n^{(m)}\right\|_\bH^2 =\frac{1}{M}\sum_{m=1}^M\text{Bias}\left(\bar w^{(m)}_{\frac{n}{M}}\right).
	\end{align}
Taking the bound of the local Bias-term $\text{Bias}\left(\bar w^{(m)}_{\frac{n}{M}}\right)$ from \ref{prop-upperbound}, proves the claim.\\
\\
$b)$ First we split the expectation operator as follows
\begin{align}
\label{varbounds}
&\mathbb{E}\left[\vv_M \otimes  \vv_M \right] \nonumber \\
&=\frac{1}{M^2}\sum_{m,m'=1}^{M}\mathbb{E}\left[\bar v^{(m)}_n \otimes \bar v^{(m')}_n \right]  \nonumber \\
&=\frac{1}{M^2}\sum_{m=1}^{M}\mathbb{E}\left[\bar v^{(m)}_n \otimes \bar v^{(m)}_n \right]+\frac{1}{M^2}\sum_{m\neq m'}\mathbb{E}\left[\bar v^{(m)}_n \otimes \bar v^{(m')}_n \right] \nonumber \\
&=:I_1+I_2.
\end{align}
Now we prove that the second operator $I_2$ is equal zero. First rewrite $I_2$ as
    
 \begin{align*}
I_2=\frac{1}{M^2}\sum_{m\neq m'}\frac{M^2}{n^2}\sum_{s,t=0}^{\frac{n}{M}-1}\mathbb{E}[ v^{(m)}_t \otimes v^{(m')}_s ].
\end{align*}
  Therefore it is enough to to prove $\mathbb{E}[ v^{(m)}_t \otimes v^{(m')}_s ]=0$ for any $m\neq m'$. Since we assume our data sets to be independent we have
$\mathbb{E}[ v^{(m)}_t \otimes v^{(m')}_s ]=\mathbb{E}[\langle . ,v^{(m)}_t \rangle]\mathbb{E}[v^{(m')}_s ]=0$, where the last equality follows from (\ref{eq:varit}).
This proves $I_2=0$. To sum up we have from (\ref{varbounds}) for the variance term,
\begin{align}
\label{eq:var-upper-uniform}
\text{Var}(\w_M)=& \frac{1}{2}\inner{ \bH , \mathbb{E}\left[\vv_M \otimes \vv_M \right] }_{op} \no \\
=&  \frac{1}{M^2}\sum_{m=1}^{M}\frac{1}{2}\inner{ \bH ,\mathbb{E}\left[\bar v^{(m)}_n \otimes \bar v^{(m)}_n \right] }_{op} \no \\
=&  \frac{1}{M^2}\sum_{m=1}^{M} \text{Var}\left(\bar w^{(m)}_{\frac{n}{M}}\right)\;.
\end{align}
Using the bound of the local variance term from \ref{prop-upperbound} completes the proof.
\end{proof}

\begin{proof}[Proof of Theorem  \ref{theo-main}]
Using lemma \ref{lem:bias-variance-deco} $a)$ we have
\[ \mbe\left[ L(\w_M) \right] - L(w^*)  \leq 2\bias(\w_M)  +  2\var(\w_M).\] 
The claim now follows from lemma \ref{distupbound}.
\end{proof}

%%%%%%%%%%%%%%%%%%%%%%%%%%%%%%%%%%%%%%%%%%%%%%%%%%%%%%%%%%%%%%%%%%%%%%%%%%%%%%%%%%%%%%%%%%%%%%%%
%%%%%%% PROOFS LOWER BOUND
%%%%%%%%%%%%%%%%%%%%%%%%%%%%%%%%%%%%%%%%%%%%%%%%%%%%%%%%%%%%%%%%%%%%%%%%%%%%%%%%%%%%%%%%%%%%%%%%

\subsection{Proofs Lower Bound }
\label{supp:proofsB}

%%%%%%%%%%%%%%%%%%%%%%%%%%%%%%%%%%%%%%%%%%%%%%%%%%%%%%%%%%%%%%%%%%%%%%%%
%%% Lower Bound Bias
%%%%%%%%%%%%%%%%%%%%%%%%%%%%%%%%%%%%%%%%%%%%%%%%%%%%%%%%%%%%%%%%%%%%%%%%

\subsubsection{Lower Bound Bias}

\begin{proposition}[Lower Bound Bias]
\label{prop:lower-bias}
Suppose Assumptions \ref{ass:second-moment} and \ref{ass:moment-lower} are satisfied and let $\gamma < \frac{1}{||\bH||}$. Recall the definition of $\bias (\w_M)$
in Lemma \ref{lem:bias-variance-deco}. The bias of the distributed SGD estimator satisfies the lower bound 
\[ \bias (\w_M) \geq  \frac{M(M-1)}{100 \gamma^2 n^2}
\left(   ||w_1 - w^*||^2_{\bH^\dagger_{0:k^*}}  + \frac{\gamma^2 n^2}{M^2}||w_1 - w^*||^2_{\bH_{k^*:\infty}} \right) \;. \]
\end{proposition}

\vspace{0.2cm}

\begin{proof}[Proof of Proposition \ref{prop:lower-bias}]
From the definition of the bias in  Lemma \ref{lem:bias-variance-deco}, we have 
\begin{align}
\label{eq:lower-b1}
\bias (\w_M) &= \frac{1}{2}\inner{ \bH , \mbe\left[ \bb_M \otimes \bb_M \right] }_{op} \nonumber \\
&= \frac{1}{2M^2}\sum_{m_1=1}^M \sum_{m_2=1}^M \inner{ \bH , \mbe\left[ \bar b^{(m_1)}_n \otimes \bar b^{(m_2)}_n \right] }_{op} \nonumber  \\ 
&= \frac{1}{2M^2}\sum_{m=1}^M  \inner{ \bH , \mbe\left[ \bar b^{(m)}_n \otimes \bar b^{(m)}_n \right] }_{op} + 
\frac{1}{2M^2}\sum_{m_1\not =m_2}^M  \inner{ \bH , \mbe\left[ \bar b^{(m_1)}_n \otimes \bar b^{(m_2)}_n \right] }_{op} \;.
\end{align}
We show that the first term in the above decomposition can be lower bounded by zero. Indeed, 
from (C.2) and (C.4) in \cite{zou2021benign} we have for all $m=1, ..., M$ the local lower bound 
\begin{align*}
 \inner{ \bH , \mbe\left[ \bar b^{(m)}_n \otimes \bar b^{(m)}_n \right] }_{op} 
&\geq \frac{M^2}{n^2} \sum_{t=1}^{\frac{n}{M}}\sum_{k=t}^{\frac{n}{M}} \inner{ ( \bI-\gamma \bH)^{k-t} \bH ,  \mbe\left[ b^{(m)}_t \otimes b^{(m)}_t \right]}_{op} \\
&\geq \frac{M^2}{\gamma n^2} \inner{\bI - ( \bI-\gamma \bH)^{\frac{n}{2M}}  , \bS^{(m)}_{\frac{n}{2M}} }_{op} \;,
\end{align*}
where we set 
\[ \bS^{(m)}_{\frac{n}{2M}} := \sum_{t=1}^{\frac{n}{2M}}   \mbe\left[ b^{(m)}_t \otimes b^{(m)}_t \right]  \;.\]
Setting $\bB_1 = b^{(m)}_1 \otimes b^{(m)}_1 = (w_1 - w^*) \otimes (w_1 - w^*)$ and applying Lemma C.4 from \cite{zou2021benign} gives then for all $m=1, ..., M$ 
\begin{align*}
\bS^{(m)}_{\frac{n}{2M}} \; 
&\succeq \; \underbrace{ \frac{\theta}{4}  \tr\left[  \left( \bI - (\bI - \gamma \bH)^{\frac{n}{2M}}\right) \bB_1 \right] \cdot 
\left(  \left( \bI - (\bI - \gamma \bH)^{\frac{n}{2M}}\right) \right) }_{PSD}
+ \underbrace{\sum_{t=1}^{\frac{n}{M}} ( \bI - \gamma \bH )^t \cdot \bB_1 \cdot (\bI - \gamma \bH )^t}_{PSD} \\
&\succeq 0  \;.
%\;  \frac{\theta}{4}  
%\tr\left[  \left( \bI - (\bI - \gamma \bH)^{\frac{n}{2M}}\right) \bB_1 \right] \cdot 
%\left(  \left( \bI - (\bI - \gamma \bH)^{\frac{n}{2M}}\right) \right)   \;.
\end{align*}
Hence, %following the proof of Lemma C.5 in \cite{zou2021benign} (adapted to our local setting) gives 
\begin{align}
\label{eq:I1}
\frac{1}{2M^2}\sum_{m=1}^M  \inner{ \bH , \mbe\left[ \bar b^{(m)}_n \otimes \bar b^{(m)}_n \right] }_{op} 
&\geq \frac{1}{2M^2}\cdot \frac{M^2}{\gamma n^2} \; \sum_{m=1}^M \inner{\bI - ( \bI-\gamma \bH)^{\frac{n}{2M}}  , \bS^{(m)}_{\frac{n}{2M}} }_{op}  \nonumber \\
&\geq 0 \;.
%\frac{\theta}{8M} \frac{M^2}{\gamma n^2} \cdot \tr\left[  \left( \bI - (\bI - \gamma \bH)^{\frac{n}{2M}}\right) \bB_1 \right] 
%  \cdot  \inner{\bI - ( \bI-\gamma \bH)^{\frac{n}{2M}}  ,  \left( \bI - (\bI - \gamma \bH)^{\frac{n}{2M}}\right) }_{op}  \nonumber  \\
%&\geq \frac{1}{M}\cdot  \frac{\theta M^2}{1000 \gamma n^2 } \; 
%\left( ||w_1 - w^*||^2_{\bI_{0:k^*}}  + \frac{\gamma  n}{M}  ||w_1 -  w^*||^2_{\bH_{k^*:\infty}} \right) \; \cdot \nonumber \\
%& \;\;\;\;  \cdot \;    \left(  k^* + \frac{\gamma^2 n^2}{M^2} \sum_{j=k^*+1}^\infty \lam_j^2 \right)  \nonumber  \\
%&=   \frac{\theta M  }{1000 \gamma n } \; 
%\left( ||w_1 - w^*||^2_{\bI_{0:k^*}}  + \frac{\gamma  n}{M}  ||w_1 -  w^*||^2_{\bH_{k^*:\infty}} \right) 
%\cdot   V_{k^*}(n,M) \;,
\end{align}
%where $k^*=\max\{ k: \lam_k \geq M/(\gamma n) \}$ and 
%\[  V_{k^*}(n,M) := \frac{k^*}{n} + \gamma^2 \frac{n}{M^2} \sum_{j=k^*+1}^\infty \lam_j^2   \;.\]

\vspace{0.2cm}

We now bound the second term in \eqref{eq:lower-b1}. 
Note that by independence of the local nodes and with \eqref{eq:expec-bias} we may write for any fixed $m_1 \not = m_2$
\begin{align*}
\mbe\left[ \bar b^{(m_1)}_n \otimes \bar b^{(m_2)}_n \right] &= 
\frac{M^2}{n^2}\sum^{ \frac{n}{M}}_{t=1}\sum_{k=1}^{\frac{n}{M}} \mbe\left[ b^{(m_1)}_t\right] \otimes \mbe\left[  b^{(m_2)}_k \right] \\
&= \frac{M^2}{n^2}\sum^{ \frac{n}{M}}_{t=1}\sum_{k=1}^{\frac{n}{M}}    ( \bI - \gamma \bH )^t \cdot \bB_1 \cdot (\bI - \gamma \bH )^k \;.
\end{align*}
Hence, 
\begin{align*}
\frac{1}{2M^2}\sum_{m_1\not =m_2}^M  \inner{ \bH , \mbe\left[ \bar b^{(m_1)}_n \otimes \bar b^{(m_2)}_n \right] }_{op} 
&= \frac{1}{2M^2} \frac{M^2}{n^2}\sum_{m_1\not =m_2}^M\sum^{ \frac{n}{M}}_{t=1}\sum_{k=1}^{\frac{n}{M}} 
 \inner{ \bH ,  ( \bI - \gamma \bH )^t \cdot \bB_1 \cdot (\bI - \gamma \bH )^k }_{op} \\
&= \frac{M(M-1)}{2\gamma n^2} 
 \inner{ \sum_{k=1}^{\frac{n}{M}}(\bI - \gamma \bH )^k  \left( \bI -( \bI - \gamma \bH )^{\frac{n}{M}+1}\right),   \bB_1 }_{op} \\
&=  \frac{M(M-1)}{2\gamma^2 n^2} 
\inner{  \left( \bI -( \bI - \gamma \bH )^{\frac{n}{M}+1} \right)^2 \bH^{-1},   \bB_1 }_{op}\;.
\end{align*}
%To proceed we let $(v_j)_j$ denote the eigenvectors of the Hessian $\bH$ and set $\omega_j = \inner{v_j , w_1 - w^*}$. We then expand 
Following now the lines of the proof of Lemma C.5 in \cite{zou2021benign} (adapted to our local setting) gives 
\begin{align*}
\frac{1}{2M^2}\sum_{m_1\not =m_2}^M  \inner{ \bH , \mbe\left[ \bar b^{(m_1)}_n \otimes \bar b^{(m_2)}_n \right] }_{op} 
\geq \frac{M(M-1)}{100 \gamma^2 n^2}
\left(   ||w_1 - w^*||^2_{\bH^\dagger_{0:k^*}}  + \frac{\gamma^2 n^2}{M^2}||w_1 - w^*||^2_{\bH_{k^*:\infty}} \right) \;.
\end{align*}
Combining now the last bound with \eqref{eq:I1} and \eqref{eq:lower-b1} finally gives  
\begin{align*}
\bias (\w_M) &\geq 
%\frac{\theta M  }{1000 \gamma n } \; 
%\left( ||w_1 - w^*||^2_{\bI_{0:k^*}}  + \frac{\gamma  n}{M}  ||w_1 -  w^*||^2_{\bH_{k^*:\infty}} \right) 
%\cdot   V_{k^*}(n,M) \; +   \\
 \frac{M(M-1)}{100 \gamma^2 n^2}
\left(   ||w_1 - w^*||^2_{\bH^\dagger_{0:k^*}}  + \frac{\gamma^2 n^2}{M^2}||w_1 - w^*||^2_{\bH_{k^*:\infty}} \right) \;.
\end{align*}
\end{proof}

%%%%%%%%%%%%%%%%%%%%%%%%%%%%%%%%%%%%%%%%%%%%%%%%%%%%%%%%%%%%%%%%%%%%%%%%
%%% Lower Bound Variance
%%%%%%%%%%%%%%%%%%%%%%%%%%%%%%%%%%%%%%%%%%%%%%%%%%%%%%%%%%%%%%%%%%%%%%%%

\subsubsection{Lower Bound Variance}

\begin{proposition}[Lower Bound Variance]
\label{prop:lower-var}
Suppose Assumptions \ref{ass:second-moment} and \ref{ass:moment-lower} are satisfied and let $\frac{n}{M}\geq 500$, $\gamma < \frac{1}{||\bH||}$. 
Recall the definition of $\var (\w_M)$
in Lemma \ref{lem:bias-variance-deco}. The variance of the distributed SGD estimator satisfies the lower bound 
\[ \var (\w_M) \geq \frac{\sigma^2_{noise}}{100}  \cdot \left( \frac{k^*}{n}  + \frac{\gamma^2 n}{M^2} \sum_{j>k^*} \lam_j^2  \right)  \;. \]
\end{proposition}

\vspace{0.2cm}

\begin{proof}[Proof of Proposition \ref{prop:lower-var}]
From the definition of the variance in  Lemma \ref{lem:bias-variance-deco}, we have 
\begin{align}
\label{eq:lower-v1}
\var (\w_M) &= \frac{1}{2}\inner{ \bH , \mbe\left[ \vv_M \otimes \vv_M \right] }_{op} \nonumber \\
&= \frac{1}{2M^2}\sum_{m_1=1}^M \sum_{m_2=1}^M \inner{ \bH , \mbe\left[ \bar v^{(m_1)}_n \otimes \bar v^{(m_2)}_n \right] }_{op} \nonumber  \\ 
&= \frac{1}{2M^2}\sum_{m=1}^M  \inner{ \bH , \mbe\left[ \bar v^{(m)}_n \otimes \bar v^{(m)}_n \right] }_{op} + 
\frac{1}{2M^2}\sum_{m_1\not =m_2}^M  \inner{ \bH , \mbe\left[ \bar v^{(m_1)}_n \otimes \bar v^{(m_2)}_n \right] }_{op} \;.
\end{align}
We first lower bound the first term. By Eq. (C.3) and Lemma C.3 in \cite{zou2021benign} (adapted to our local setting) we obtain 
\begin{align*}
\frac{1}{2M^2}\sum_{m=1}^M  \inner{ \bH , \mbe\left[ \bar v^{(m)}_n \otimes \bar v^{(m)}_n \right] }_{op} 
&\geq \frac{1}{2M^2} \frac{M^2}{n^2} \sum_{m=1}^M \sum_{t=0}^{\frac{n}{M}-1} \sum_{k=t}^{\frac{n}{M}-1} 
\inner{ \left( \bI - \gamma \bH \right)^{k-t} \bH , \mbe\left[ v^{(m)}_t \otimes v^{(m)}_t  \right]  }_{op} \\ 
&\geq  \frac{\sigma^2_{noise}}{100M^2}  \sum_{m=1}^M \left( \frac{M}{n} k^* + \frac{\gamma^2 n}{M} \sum_{j>k^*} \lam_j^2  \right) \\
&= \frac{\sigma^2_{noise}}{100}   V_{k^*}(n,M) \;,
\end{align*}
where 
\[ V_{k^*}(n,M) := \left( \frac{k^*}{n}  + \frac{\gamma^2 n}{M^2} \sum_{j>k^*} \lam_j^2  \right) \;. \]
To derive the final bound we argue that the second term in \eqref{eq:lower-v1} is  zero. Indeed, by independence of the local nodes we may write 
for any $m_1 \not = m_2$ with \eqref{eq:varit} 
\begin{align*}
 \mbe\left[ v^{(m_1)}_t \otimes v^{(m_2)}_k  \right] &= \mbe\left[ v^{(m_1)}_t \right]  \otimes \mbe\left[v^{(m_2)}_k  \right] \\
 &= ( \bI - \gamma \bH )^t (v^{(m_1)}_0 \otimes v^{(m_2)}_0) (\bI - \gamma \bH  )^k \\
 &= 0 \;,
\end{align*}
since $v^{(m)}_0 =0$ for all $m=1,...,M$. Hence, 
\[ \frac{1}{2M^2}\sum_{m_1\not =m_2}^M  \inner{ \bH , \mbe\left[ \bar v^{(m_1)}_n \otimes \bar v^{(m_2)}_n \right] }_{op} = 0 \;. \]
this finishes the proof. 
\end{proof}

%%%%%%%%%%%%%%%%%%%%%%%%%%%%%%%%%%%%%%%%%%%%%%%%%%%%%%%%%%%%%%%%%%%%%%%%%%%%%%%%%%%%%%%%%%%%%%%%
%%%%%%%%%%%%%%%% PROOFS RATE OF CONVERGENCE
%%%%%%%%%%%%%%%%%%%%%%%%%%%%%%%%%%%%%%%%%%%%%%%%%%%%%%%%%%%%%%%%%%%%%%%%%%%%%%%%%%%%%%%%%%%%%%%%

\subsection{Proofs Rates of Convergence}
\label{supp:proofsC}

\begin{proof}[Proof of Corollary \ref{cor:spiked}]
Let the sequence $M_n \leq \sqrt{\frac{\gamma(1-2\gamma \tau) n }{ R^2}} $.  
By definition of $k^*$ we know that $k^*= \tilde d = \left( \frac{n}{M_n} \right)^r$ and hence $\lam_{k^*} = \left( \frac{M_n}{n} \right)^r$. We first bound the bias from 
Theorem \ref{theo-main}. Since $||w^*||_2 \leq R$ by assumption, we find 
\begin{align}
||w^*||^2_{\bH^\dagger_{0:k^*}} &\leq  \frac{||w^*||^2_2}{\lam_{k^*}} \leq R^2\left( \frac{n}{M_n} \right)^r \;. \label{1.term}
\end{align}
Similarly, since $\frac{n}{M_n} \to \infty$ as $n \to \infty$, there exists $ n_0 \in \mbn$ such that  
\begin{align}
||w^*||^2_{\bH_{k^*:\infty}} \leq \frac{R^2}{ \left( \frac{n}{M_n} \right)^q - \left( \frac{n}{M_n} \right)^r } \leq c_{n_0} R^2\left( \frac{M_n}{n} \right)^q \;, \label{2.term}
\end{align}
for any $n \geq n_0$ and some $c_{n_0} < \infty$. 
Using that $\tr[\bH] = 2$ and $||w^*||^2_{\bI_{0:k^*}} \leq R^2$, we find for all $n \geq n_0$, for some $n_0 \in \mbn$, that 
\begin{align*}
 \frac{2\tau M^2\left(  ||w^*|^2_{\bI_{0:k}} + \gamma \frac{n}{M} || w^*||^2_{\bH_{k:\infty}} \right) }{\gamma  n (1-\gamma \tau \tr[\bH ]) } 
 &\leq 4  \max \{ 1, c_{n_0}\} \; \frac{\tau R^2}{1-2\gamma \tau} \; \frac{M_n^2}{\gamma n} \;.
\end{align*}
Note that we also use that $M_n \leq n$ and hence $\left( \frac{M_n}{n}\right)^{q-1} \leq 1$, since $q >1$. Since 
\[ M_n \leq \sqrt{\frac{\gamma(1-2\gamma \tau) n }{ R^2}} \]
we have 
\[ \frac{\tau R^2}{1-2\gamma \tau} \; \frac{M_n^2}{\gamma n} \leq 1  \]
and hence 
\begin{align}
 \frac{2\tau M_n^2\left(  ||w^*|^2_{\bI_{0:k}} + \gamma \frac{n}{M} || w^*||^2_{\bH_{k:\infty}} \right) }{\gamma  n (1-\gamma \tau \tr[\bH ]) } 
 &\leq 4  \max \{ 1, c_{n_0}\} \;. \label{3.term}
\end{align}
We further observe that by the definition of the spectrum of $\bH$
\[ \sum_{j > k^*} \lam_l^2  = \sum_{j=\tilde d}^d \frac{1}{d - \tilde d} = \frac{1}{ \left( \frac{n}{M_n} \right)^q - \left( \frac{n}{M_n} \right)^r } 
\leq c_{n_0} \left( \frac{M_n}{n} \right)^q \;,\]
for any $n$ sufficiently large, by using the argumentation as above. Hence, 
\begin{align}
\label{eq:V}
V_{k^*}(n,M_n) 
&:= \frac{k^*}{n} + \gamma^2 \frac{n}{M_n^2} \sum_{j=k^*+1}^\infty \lam_j^2 \nonumber \\
&\leq  \max \{ 1, c_{n_0}\} \; \frac{1}{M_n} \cdot \left( \; \left(  \frac{M_n}{n} \right)^{1-r} + \gamma^2 \left( \frac{M_n}{n} \right)^{q-1} \;\right) \;.
\end{align}
Combining (\ref{1.term}), (\ref{2.term}), (\ref{3.term}) and (\ref{eq:V}),  we find for the bias term 
\begin{align}
\label{eq:bias-1}
 \bias (M_n) &\leq  \frac{R^2}{\gamma^2 (n/M_n)^2}\left( \frac{n}{M_n} \right)^r + c_{n_0} R^2\left( \frac{M_n}{n} \right)^q  + 4  \max \{ 1, c_{n_0}\}V_{k^*}(n,M_n)  \nonumber \\
 &\leq \max \{1, c_{n_0} \} \; R^2 \left(  \frac{1}{\gamma^2} \left( \frac{M_n}{n} \right)^{2-r} + \left( \frac{M_n}{n} \right)^q  \right)\;+ \\
&\hspace{0.5cm}4\max \{ 1, c_{n_0}\}^2 \; \frac{1}{M_n} \cdot \left( \; \left(  \frac{M_n}{n} \right)^{1-r} + \gamma^2 \left( \frac{M_n}{n} \right)^{q-1} \;\right) \;.  
\end{align}
We now turn to the bound of the variance term. 
From \eqref{eq:V} we have 
\begin{align*}
 \var(M_n)  \leq  \max \{ 1, c_{n_0}\}\left( \frac{\sigma^2}{1-\gamma \tau \tr[\bH]}\right) 
\cdot \frac{1}{M_n} \cdot \left( \; \left(  \frac{M_n}{n} \right)^{1-r} + \gamma^2 \left( \frac{M_n}{n} \right)^{q-1} \;\right) \;.
\end{align*}

Combining the bounds for bias and variance leads to the total error bound
\begin{align*}
& \mbe\left[ L(\w_{M_n}) \right] - L(w^*)  \leq    \\
& 2\tilde c_{n_0} \cdot c_{\gamma, \tau , \sigma}\cdot 
\left(  R^2 \left(  \frac{1}{\gamma^2} \left( \frac{M_n}{n} \right)^{2-r} + \left( \frac{M_n}{n} \right)^q  \right)  + 
 \cdot \frac{1}{M_n} \cdot \left( \left(  \frac{M_n}{n} \right)^{1-r} + \gamma^2 \left( \frac{M_n}{n} \right)^{q-1} \right)   \right) \;,
\end{align*}
with 
\[  c_{\gamma, \tau , \sigma} :=  1+\frac{\sigma^2}{1-\gamma \tau \tr[\bH]} \;, \quad \tilde c_{n_0} = 4\max \{ 1, c_{n_0}\}^2\;,\]
holding for any $n$ sufficiently large. We proceed by further simplifying the right hand side of the above inequality. Since $\tau \geq 1$ and $1-\gamma \tau \tr[\bH] < 1$, 
the assumption on $M_n$ implies that 
\[ M_n^2 \leq \frac{n \gamma}{R^2}  \;,\]
further implying that 
\[  \frac{R^2}{\gamma} \left( \frac{M_n}{n} \right)^{2-r} \leq \frac{1}{M_n} \left(  \frac{M_n}{n} \right)^{1-r} \]
and 
\[ R^2  \left( \frac{M_n}{n} \right)^q  \leq \frac{\gamma }{M_n} \left( \frac{M_n}{n} \right)^{q-1} \;. \]
As a result, applying Theorem \ref{theo-main}, the excess risk can be bounded by 
\begin{align*}
 \mbe\left[ L(\w_{M_n}) \right] - L(w^*) & \leq    
  4\tilde c_{n_0} \cdot c_{\gamma, \tau , \sigma}\cdot \frac{1}{M_n} 
\left( \left(\frac{1}{\gamma} + 1\right) \left(  \frac{M_n}{n} \right)^{1-r} + (\gamma + \gamma^2) \left( \frac{M_n}{n} \right)^{q-1}   \right) \\
&\leq  4\tilde c_{n_0} \cdot c_{\gamma, \tau , \sigma}\cdot \frac{1}{\gamma M_n} 
\left(  \left(  \frac{M_n}{n} \right)^{1-r} + \left( \frac{M_n}{n} \right)^{q-1}   \right) \;.
\end{align*}
In the last step we use that $\gamma < \frac{1}{2\tau} < \frac{1}{2}$.
\end{proof}

\vspace{0.3cm}

\begin{proof}[Proof of Corollary \ref{cor:poly}]
Assume the sequence $(M_n)_n$ satisfies $M_n/n \to 0$ as $n \to \infty$.  
We use Theorem \ref{theo-main} to bound the excess risk and find estimates for bias and variance. By the definition of $k^*$ we have 
\[ k^* = \max \left\{ k \in \mbn : k\leq \left( \frac{\gamma n}{M_n} \right)^{\frac{1}{1+r}}  \right\} = 
\left\lfloor \left( \frac{\gamma n}{M_n} \right)^{\frac{1}{1+r}} \right\rfloor \;. \]
Hence, there exists $n_0 \in \mbn$ such that for all $n \geq n_0$
\[  c_{n_0} \left( \frac{\gamma n}{M_n} \right)^{\frac{1}{1+r}}  \leq k^* \leq C_{n_0}\left( \frac{\gamma n}{M_n} \right)^{\frac{1}{1+r}} \;, \]
for some constants $0 < c_{n_0} \leq C_{n_0}$. Therefore, 
\[  \lam_{k^*} = (k^*)^{-(1+r)} \leq \left( \frac{1}{c_{n_0}}\right)^{1+r} \cdot \frac{M_n}{\gamma n}  \]
and 
\[ \frac{1}{ \lam_{k^*}} = (k^*)^{1+r} \leq C^{1+r}_{n_0} \cdot \frac{n}{\gamma M_n} \;. \]
We therefore get for the first two terms of the bias 
\begin{align}
\frac{M_n^2}{\gamma^2 n^2}\cdot  ||w^*||^2_{\bH^\dagger_{0:k^*}} &\leq \frac{R^2M_n^2}{\gamma^2 n^2 \lam_{k^*}} \\
&\leq C_{n_0}^{1+r} \frac{R^2}{\gamma} \cdot \frac{M_n}{n} \;.\label{term-1}
\end{align}
and 
\begin{align}
||w^*||^2_{\bH_{k^*:\infty}}\leq R^2 \lam_{k^*} \leq R^2 \left( \frac{1}{c_{n_0}}\right)^{1+r} \cdot \frac{M_n}{\gamma n} \;. \label{term-2}
\end{align}
We now bound the last term of the bias. To this end, we apply a well known bound for sums over decreasing functions, i.e., 
\[ \sum_{j \geq k} f(j) \leq \int_k^\infty f(x) dx \;. \]
This gives 
\[ \sum_{j > k^*} \lam_j^2 \leq \int_{k^*}^\infty x^{-2(r+1)} dx \leq \frac{1}{2r+1} (k^*)^{-(2r+1)} \leq  
\frac{1}{2r+1} c_{n_0}^{-(2r+1)} \left( \frac{M_n}{\gamma n} \right)^{1+\frac{r}{1+r}} \;. \]
Thus, 
\begin{align}
\label{eq:v2}
  V_{k^*}(n,M_n) &=  \frac{k^*}{n} + \gamma^2 \frac{n}{M^2} \sum_{j=k^*+1}^\infty \lam_j^2  \nonumber  \\
                 &\leq \frac{1}{n} C_{n_0}\left( \frac{\gamma n}{M_n} \right)^{\frac{1}{1+r}} + 
\gamma^2 \frac{n}{M_n^2} \frac{c_{n_0}^{-(2r+1)} }{2r+1} \left( \frac{M_n}{\gamma n} \right)^{1+\frac{r}{1+r}}  \nonumber  \\
&\leq c'_{r, n_0} \left( \frac{1}{n} \left( \frac{\gamma n}{M_n} \right)^{\frac{1}{1+r}} + 
 \frac{\gamma }{M_n}  \left( \frac{M_n}{\gamma n} \right)^{\frac{r}{1+r}}  \right)  \nonumber  \\
&\leq 2  c'_{r, n_0} \cdot \frac{\gamma }{M_n}  \left( \frac{M_n}{\gamma n} \right)^{\frac{r}{1+r}}   \;,
\end{align}  
with  
\[   c'_{r, n_0} = \max\left\{  C_{n_0} , \frac{c_{n_0}^{-(2r+1)} }{2r+1} \right\} \;. \]
Moreover,  
\begin{align*}
  \frac{2\tau M_n^2\left(  ||w^*||^2_{\bI_{0:k^*}} + \gamma \frac{n}{M_n} || w^*||^2_{\bH_{k^*:\infty}} \right) }{\gamma  n (1-\gamma \tau \tr[\bH ]) } 
 &\leq \frac{2\tau M_n^2}{\gamma  n (1-\gamma \tau \tr[\bH ])} \left( R^2 + R^2 \gamma \frac{n}{M_n} \lam_{k^*} \right) \\
 &\leq c''_{n_0}\;  \frac{2\tau M_n^2}{\gamma  n (1-\gamma \tau \tr[\bH ])} R^2 \left( 1 + \gamma \frac{n}{M_n} \frac{M_n}{\gamma n}  \right) \\
 &\leq 2c''_{n_0}\;  \frac{2\tau}{ (1-\gamma \tau \tr[\bH ])} \cdot \frac{ R^2  M_n^2}{\gamma  n} \;,
\end{align*}
with
\[ c''_{n_0}=    \max\left\{ 1 ,  \left( \frac{1}{c_{n_0}}\right)^{1+r} \right\} \;.  \]
Hence, combining this with \eqref{eq:v2} and choosing 
\begin{equation}
\label{eq:M2}
 M_n \leq \frac{\sqrt{\gamma n}}{R} 
\end{equation} 
leads to 
\begin{align}
%\label{eq:v3}
& \frac{2\tau M_n^2\left(  ||w^*||^2_{\bI_{0:k^*}} + \gamma \frac{n}{M_n} || w^*||^2_{\bH_{k^*:\infty}} \right) }{\gamma  n (1-\gamma \tau \tr[\bH ]) } \cdot  V_{k^*}(n,M_n)\nonumber \\
&\leq 2c''_{n_0}\;  \frac{2\tau}{ (1-\gamma \tau \tr[\bH ])} \cdot \frac{ R^2  M_n^2}{\gamma  n}  \cdot 
2  c'_{r, n_0} \cdot \frac{\gamma }{M_n}  \left( \frac{M_n}{\gamma n} \right)^{\frac{r}{1+r}} \nonumber \\
&\leq c_{r, n_0}\frac{\tau}{ (1-\gamma \tau \tr[\bH ])} \cdot
 \frac{ R^2  M_n^2}{\gamma  n} \cdot \frac{\gamma }{M_n}  \left( \frac{M_n}{\gamma n} \right)^{\frac{r}{1+r}} \nonumber \\
&\leq c_{r, n_0}\frac{\tau}{ (1-\gamma \tau \tr[\bH ])} \cdot   \frac{\gamma}{M_n} \left( \frac{M_n}{\gamma n} \right)^{\frac{r}{1+r}}\;,\label{term-3}
\end{align}  
where $c_{r, n_0}=8 c''_{n_0}\cdot c'_{r, n_0}$.
Combining (\ref{term-1}), (\ref{term-2}) and (\ref{term-3}), we find for all $n \geq n_0$
\begin{equation}
\label{eq:bias-poly}
 \bias(M_n) \leq  \tilde c_{r, n_0} \cdot \frac{R^2}{\gamma} \cdot  \frac{M_n}{n}  + c_{r, n_0}\frac{\tau}{ (1-\gamma \tau \tr[\bH ])} \cdot   \frac{\gamma}{M_n} \left( \frac{M_n}{\gamma n} \right)^{\frac{r}{1+r}}\;, 
\end{equation} 
where we set 
\[ \tilde c_{r, n_0}= 2\max\left\{  C_{n_0}^{1+r} , \left( \frac{1}{c_{n_0}}\right)^{1+r}  \right\}  \;.\]
We now turn to bounding the variance. Using \eqref{eq:v2}  once more, the variance can be bounded by 
\begin{align*}
 \var(M_n) \leq  \frac{\sigma^2}{1-\gamma \tau \tr[\bH]} \cdot 2  c'_{r, n_0} \cdot \frac{\gamma }{M_n}  \left( \frac{M_n}{\gamma n} \right)^{\frac{r}{1+r}}   \;.
\end{align*} 
Combining the bias bound \eqref{eq:bias-poly} with the variance bound, we obtain for the excess risk 
\begin{align*}
 \mbe\left[ L(\w_{M_n}) \right] - L(w^*) & \leq c_{r,n_0, \tau, \sigma } \left( \frac{R^2}{\gamma} \cdot  \frac{M_n}{n} + 
 \frac{\gamma }{M_n}  \left( \frac{M_n}{\gamma n} \right)^{\frac{r}{1+r}} \right) \;.
\end{align*} 
\[  c_{r,n_0, \tau, \sigma }: =   \max\left\{   \tilde c_{r, n_0}  ,  2 c_{r, n_0} \cdot  c'_{r, n_0} \cdot \frac{\max\{ \tau , \sigma^2\}}{ 1-\gamma \tau \tr[\bH ]} \right\} \;. \]
Note that the choice 
\begin{equation}
\label{eq:M3}
  M_n \leq \left(\frac{\gamma }{R^2}\right)^{\frac{1+r}{2+r}} \cdot (\gamma n)^{\frac{1}{2+r}} 
\end{equation}  
leads to a dominating variance part, i.e. 
\[   \frac{R^2}{\gamma} \cdot  \frac{M_n}{n} \leq \frac{\gamma }{M_n}  \left( \frac{M_n}{\gamma n} \right)^{\frac{r}{1+r}} \]
and 
\begin{align*}
 \mbe\left[ L(\w_{M_n}) \right] - L(w^*) & \leq 2c_{r,n_0, \tau, \sigma }  \frac{\gamma }{M_n}  \left( \frac{M_n}{\gamma n} \right)^{\frac{r}{1+r}}  \;.
\end{align*} 
Note that the choice \eqref{eq:M3} is compatible with the choice \eqref{eq:M2}, i.e., 
\[   M_n \leq \left(\frac{\gamma }{R^2}\right)^{\frac{1+r}{2+r}} \cdot (\gamma n)^{\frac{1}{2+r}} \leq \frac{\sqrt{\gamma n}}{R} \;, \]
following from the fact that $r >0$, provided that $n$ is sufficiently large.
\end{proof}

%%%%%%%%%%%%%%%%%%%%%%%%%%%%%%%%%%%%%%%%%%%%%%%%%%%%%%%%%%%%%%%%%%%%%%%%%%%%%%%%%%%%%%%%%%%%%%%%%%%%%%%%%%%%
%%%%%%%%%   THE BENEFIT OF IMPLICIT REGULARIZATION     %%%%%%%%%%%%%%%%%%%%%%%%%%%%%%%%%%%%%%%%%%%%%%%%%%%%%%%%%%%%%%%%%%%%%%%%%%%%%%%%%%%%%%
%%%%%%%%%%%%%%%%%%%%%%%%%%%%%%%%%%%%%%%%%%%%%%%%%%%%%%%%%%%%%%%%%%%%%%%%%%%%%%%%%%%%%%%%%%%%%%%%%%%%%%%%%%%%

\section{PROOFS SECTION 4 ( COMPARISON OF SAMPLE COMPLEXITY OF DSGD AND DRR)}

\subsection{Lower Bound for distributed ridge regression}

In this section we derive a lower bound for the distributed RR estimator.  
We adopt the following notation and assumptions from \cite{tsigler2020benign}. 

\begin{itemize}
\item $\bH^{-1/2}x$, where $x \in \mbr^d$ is sub-Gaussian with independent components
\item $\bX=(\sqrt{\lam_1} z_1, ..., \sqrt{\lam_d}z_d) $ with $z_j$ being sub-Gaussian with independent components
\item $ \bA:=  \bX \bX^T  + \lam I_n$, \;\;\; $\bA_m := \bX_m \bX_m^T  + \lam I_n $
\item $\bA_{-j} = \sum_{i \not j} \lam_i z_i z_i^T + \lam I_n $
%\item $\bB_{ij} = \inner{\bB e_i , e_j}$
\end{itemize}

%\todo{we need $\mbe_{\epsilon_m}[\epsilon_m]  = 0$; state distributional assumptions!}

\vspace{0.3cm}

Crucial for our analysis is the following quantity, called the \emph{local effective dimension} for the RR problem: 
\begin{equation}  
\label{eq:eff-dim-RR}
k^*_{\ols } := \min \left\{ k: \lam_{k+1} \leq \frac{ M\left( \lam + \sum_{j > k} \lam_j\right)}{bn} \right\}  \;.
\end{equation}

%%%%%%%%%%%%%%%%%%%%%%%%%%%%%%%%%%%%%%%%%%%%%%%%%%%%%%%%%%%%%%%%%%%%%%%%%
%%%%%%%%%%%%%%%%%%%%%%%%%%%%%%%%%%%%%%%%%%%%%%%%%%%%%%%%%%%%%%%%%%%%%%%%%

\subsubsection{Bias-Variance Decomposition DRR}

\begin{definition}[Bias and Variance of Distributed RR]
\label{def:terms}
Let 
\[  \Pi_m(\lam) :=   \left( \bX^T_m \bX_m + \lam \right)^{-1}\bX_m^T \bX_m - Id \;,  \]
\[ \widehat{\bias} (\olsest_n (\lam ))  :=  \left| \left| \bH^{1/2} \left( \frac{1}{M} \sum_{m=1}^M \Pi_m(\lam) w^*      \right)     \right| \right| ^2 \;, \]
\[ \widehat{\var} (\olsest_n (\lam ))  := 
 \left| \left| \bH^{1/2} \left( \frac{1}{M} \sum_{m=1}^M ( \bX^T_m \bX_m + \lam )^{-1}\bX_m^T  \epsilon_m   \right)     \right| \right| ^2 \;.\] 
We call 
\begin{align*}
\bias (\olsest_n (\lam )) &= \mbe\left[ \widehat{\bias} (\olsest_n (\lam )) \right] 
\end{align*}
the (expected) bias of the distributed RR estimator and 
\begin{align*}
\var(\olsest_n  (\lam )) &=  \mbe\left[ \widehat{\var} (\olsest_n (\lam ))  \right] 
\end{align*}
the (expected) variance.
\end{definition}

\vspace{0.2cm}
We immediately obtain:
\vspace{0.2cm}

\begin{lemma}
\label{prop:risk-local-RR}
The  excess risk satisfies
 \[ \mbe\left[  || \bH^{1/2}( \olsest_n (\lam ) - w^* ) ||^2 \right] = \bias (\olsest_n (\lam )) +  
 \var(\olsest_n  (\lam ))\;.  \]
\end{lemma}

\vspace{0.2cm}

\begin{proof}[Proof of Lemma \ref{prop:risk-local-RR}]
We split the excess risk as 
\begin{align*}
|| \bH^{1/2}( \olsest_n (\lam ) - w^* ) ||^2 
&= \left| \left| \bH^{1/2} \left( \frac{1}{M} \sum_{m=1}^M \hat w_m^{\ols }(\lam ) - w^*  \right)     \right| \right|^2 \\
&= \left| \left| \bH^{1/2} \left( \frac{1}{M} \sum_{m=1}^M ( \bX^T_m \bX_m + \lam )^{-1}\bX_m^T \bY_m  - w^*  \right)     \right| \right|^2 \\
&= \left| \left| \bH^{1/2} \left( \frac{1}{M} \sum_{m=1}^M ( \bX^T_m \bX_m + \lam )^{-1}\bX_m^T  ( \bX_m w^* + \epsilon_m )   - w^*  \right)     \right| \right| ^2 \\
&= \widehat{\bias} (\olsest_n (\lam )) +  \widehat{\var} (\olsest_n (\lam ))  \\
& + \frac{2}{M^2} \sum_{m=1}^M \sum_{m'=1}^M  \inner{  \bH  \; \Pi_m(\lam)w^*  ,  ( \bX^T_m \bX_m + \lam )^{-1}\bX_m^T  \epsilon_m   } \;.
\end{align*}
We argue that the expectation with respect to the noise (i.e. conditioned on $\bX$) of the last term is equal to zero. 
Indeed, by linearity and since $\epsilon_m$ is centered (conditioned on $\bX_m$) for all $m \in [M]$, we find  
\begin{align*}
\mbe_{\epsilon_m} \left[   \inner{  \bH  \; \Pi_m(\lam)w^*  ,  ( \bX^T_m \bX_m + \lam )^{-1}\bX_m^T  \epsilon_m   } \right] 
&=  \inner{  \bH  \; \Pi_m(\lam)w^*  ,  ( \bX^T_m \bX_m + \lam )^{-1}\bX_m^T   \mbe_{\epsilon_m} \left[\epsilon_m  \right] }  \\
&=0 \;. 
\end{align*}
Hence, 
\[ \mbe\left[  || \bH^{1/2}( \olsest_n (\lam ) - w^* ) ||^2 \right] = \mbe\left[ \widehat{\bias} (\olsest_n (\lam )) \right]   +  
 \mbe\left[ \widehat{\var} (\olsest_n (\lam ))  \right] \;.  \]
\end{proof}

%%%%%%%%%%%%%%%%%%%%%%%%%%%%%%%%%%%%%%%%%%%%%%%%%%%%%%%%%%%%%%%%%%%%%%%%%%%%%%%%%%%
%%%%%%%%%%%% Bias lower Bound RR
%%%%%%%%%%%%%%%%%%%%%%%%%%%%%%%%%%%%%%%%%%%%%%%%%%%%%%%%%%%%%%%%%%%%%%%%%%%%%%%%%%%

\subsubsection{Lower Bound of Bias for DRR}

\begin{proposition}[Lower Bound of Bias for local RR]
\label{prop:low-bias-RR}
Assume $\bH$ is strictly positive definite with $\tr[H]< \infty$. There exist absolute constants $b>1$, $c>1$ such that 
\[ \bias (\olsest_n (\lam ))  \geq \frac{M-1}{c M}\cdot 
\left(   \frac{M^2 \left( \lam + \sum_{j > k^*_{\ols}} \lam_j \right)^2}{n^2} \cdot ||w^*||^2_{\bH^{-1}_{0:k^*_{\ols}}}  + 
||w^*||^2_{\bH_{k^*_{\ols} : \infty}} \right) \;, \]
where $k^*_{\ols}$ is defined in \eqref{eq:eff-dim-RR}.
\end{proposition}

\vspace{0.2cm}
For proving this Proposition we need the following Lemma. 

\begin{lemma}
\label{lem:low-bias-1}
Let $\tilde \bX \in \mbr^{n \times d}$ be an independent copy of $\bX \in \mbr^{n \times d}$ and set 
$\tilde \bA = \tilde \bX\tilde \bX^T  + \lam$. 
Define further the operator 
\[ \bB := (I_d - \bX^T \bA^{-1} \bX)\bH (I_d - \tilde \bX^T \tilde \bA^{-1} \tilde \bX) \;. \]
\begin{enumerate}
\item
For any $i\not = j$, we have  
\[ \mbe_{\bX , \tilde \bX}\left[ \bB_{ij}\right] = 0 \;. \]

\item The diagonal elements satisfy for any $k$
\[ \mbe_{\bX , \tilde \bX}\left[ \bB_{ii}\right] 
\geq \frac{1}{c}\cdot \frac{\lam_i}{\left( 1 + \frac{\lam_i}{\lam_{k+1}}\cdot \frac{n}{\rho_k} \right)^2} \;, \]
for some absolute constant $c>1$ and where we define 
\[ \rho_k = \frac{\lam + \sum_{j > k} \lam_j}{\lam_{k+1}} \;. \] 
\end{enumerate}
\end{lemma}

\vspace{0.2cm}

\begin{proof}[Proof of Lemma \ref{lem:low-bias-1}]
Recall that $\bH = diag\{ \lam_1, ..., \lam_d\}_j$ and 
\[  \bX = ( \sqrt{\lam_1}z_1 , ..., \sqrt{\lam_d} z_d ) \;, \quad  \tilde \bX = ( \sqrt{\lam_1} \tilde z_1 , ..., \sqrt{\lam_d} \tilde z_d ) \;.\]
\begin{enumerate}
\item 
Let $i \not = j$. We expand 
\begin{align*}
\bB_{ij} 
&= \underbrace{\inner{e_i , \bH e_j }}_{=0} - \sqrt{\lam_i} \inner{e_j , \bH \bX^T \bA^{-1} z_i} - \sqrt{\lam_j} \inner{ e_i , \bH \tilde X^T \tilde A^{-1} \tilde z_j  } 
 + \sqrt{z_i z_j} \inner{z_i , \bA^{-1} \bX \bH \tilde \bX^T \tilde \bA^{-1} \tilde z_j} \no \\
&= -\lam_j \sqrt{\lam_i \lam_j} \inner{z_j , \bA^{-1} z_i} - \lam_i \sqrt{\lam_i \lam_j} \inner{\tilde z_i , \tilde \bA^{-1}\tilde z_j} + 
  \sqrt{\lam_i \lam_j} \inner{ z_i , \bA^{-1} \bX \bH \tilde \bX^T \tilde \bA^{-1} \tilde z_j  } \;. 
\end{align*}
We define the map $F(z_j) := \inner{z_j , \bA^{-1} z_i}$. Following the lines of the proof of Lemma C.7 in \cite{zou2021benefits} shows that $\mbe_{z_j}[F(z_j)] = 0$. 
Using similar arguments, the same is true for the second and last term, showing the result. 
\item  
We expand 
\begin{align*}
\bB_{ii} 
&= \inner{ \bH(e_i - \sqrt{\lam_i}\bX^T \bA^{-1}z_i) , e_i - \sqrt{\lam_i} \tilde \bX^T \tilde \bA^{-1} \tilde z_i  } \no \\
&= \underbrace{\inner{\bH e_i , e_i}}_{= \lam_i} +  \lam_i \inner{\bH \bX^T \bA^{-1}z_i ,  \tilde \bX^T \tilde \bA^{-1} \tilde z_i}  
 - \sqrt{\lam_i}  \inner{\bH e_i , \tilde \bX^T \tilde \bA^{-1} \tilde z_i}   - \sqrt{\lam_i}\inner{\bH e_i ,\bX^T \bA^{-1}z_i } \no \\
&=  \lam_i\left[  1- \lam_i \left(  \inner{z_i , \bA^{-1}z_i} + \inner{\tilde z_i , \tilde  \bA^{-1}\tilde  z_i}  \right) \right] + 
\lam_i \inner{ \bH\bX^T\bA^{-1}z_i  ,\tilde \bX^T \tilde \bA^{-1} \tilde z_i  } \;.
\end{align*}
Setting 
\[  a_i:= \inner{z_i , \bA^{-1} z_i}  \;, \quad \tilde a_i:= \inner{ \tilde z_i , \tilde \bA^{-1} \tilde z_i}  \]
we further find that 
\begin{align*}
\lam_i \inner{ \bH\bX^T\bA^{-1}z_i  ,\tilde \bX^T \tilde \bA^{-1} \tilde z_i  }  
&= \lam_i \sum_{j=1}^d \lam_j (\bX^T\bA^{-1}z_i )_j \cdot (\tilde \bX^T \tilde \bA^{-1} \tilde z_i)_j \no \\
&=  \lam_i \sum_{j=1}^d \lam^2_j \inner{z_j , \bA^{-1}z_i} \cdot \inner{\tilde z_j, \tilde \bA^{-1}z_i} \no \\
&=  \lam^3_i \cdot a_i \cdot \tilde a_i + \lam_i \sum_{j \not = i} \lam_j^2 \inner{z_j , \bA^{-1}z_i} \cdot \inner{\tilde z_j, \tilde \bA^{-1}z_i}  \;.
%&= \lam_i \inner{ \bA^{-1}z_i ,  \sum_{j=1}^d \lam^2_j  \inner{ \tilde z_j , \tilde \bA^{-1}\tilde z_j } z_j   } 
\end{align*}
By independence, the last term is non-negative in expectation, i.e.
\begin{align*}
& \mbe_{\bX , \tilde\bX}\left[ \lam_i \sum_{j \not = i} \lam_j^2 \inner{z_j , \bA^{-1}z_i} \cdot \inner{\tilde z_j, \tilde \bA^{-1}z_i} \right] \\
& = \lam_i \sum_{j \not = i} \lam_j^2 \mbe_{\bX}\left[  \inner{z_j , \bA^{-1}z_i} \right] \cdot \mbe_{ \tilde\bX}\left[  \inner{\tilde z_j, \tilde \bA^{-1}z_i} \right] \\
&= \lam_i \sum_{j \not = i} \lam_j^2 \cdot \mbe_{\bX}\left[  \inner{z_j , \bA^{-1}z_i} \right]^2   \\
&\geq 0 \;.
\end{align*}
Hence, for deriving a lower bound in expectation it is sufficient to lower bound the expression 
\[  \lam_i \cdot \left[  1- \lam_i \left(  a_i +\tilde a_i  \right) \right] +  \lam^3_i \cdot a_i \cdot \tilde a_i  
   =   \lam_i \cdot   (1-\lam_i a_i)\cdot ( 1-\lam_i \tilde a_i) \;. \]
Using independence once more we find 
\begin{align*}
\mbe_{\bX , \tilde \bX}\left[ \bB_{ii}\right]  &\geq  \lam_i \cdot   \mbe_{\bX , \tilde \bX}\left[   (1-\lam_i a_i)\cdot ( 1-\lam_i \tilde a_i) \right] \;.
%&= \lam_i \cdot   \mbe_{\bX , \tilde \bX}\left[   (1-\lam_i a_i)\right]^2 \;. 
\end{align*}
We proceed as in the proof of Lemma C.7 in \cite{zou2021benefits}. Recall that 
\[ (1-\lam_i a_i) = \frac{1}{1+ \lam_i \inner{z_i , \bA_{-i}^{-1}z_i}} \]
and for all $k$
\[ \inner{z_i , \bA_{-i}^{-1}z_i} \leq c\cdot \frac{n}{\lam_{k+1} \rho_k} \;, \]
for some $c>0$, with high probability. 
Concluding as in \cite{zou2021benefits} and using independence finishes the proof.
\end{enumerate}
\end{proof}

\vspace{0.3cm}

\begin{proof}[Proof of Proposition \ref{prop:low-bias-RR}]
Setting $w_m (\lam ) = \bH^{1/2} \Pi_m (\lam ) w^*$ (see Definition \ref{def:terms}), we decompose the bias as 
\begin{align}
\label{eq:boring}
 \bias (\olsest_n (\lam ))  &= \mbe\left[ \widehat{\bias} (\olsest_n (\lam )) \right] \no \\
 &=  \mbe\left[  \left| \left|   \frac{1}{M} \sum_{m=1}^M w_m (\lam )        \right| \right|^2   \right]  \no  \\
 &= \frac{1}{M^2}  \mbe\left[  \tr\left[  \left(  \sum_{m=1}^Mw_m (\lam ) \right) \otimes \left(  \sum_{m'=1}^Mw_{m'} (\lam ) \right)   \right] \right]  \no   \\
 &= \frac{1}{M^2}  \sum_{m=1}^M   \mbe\left[   \tr\left[  w_m (\lam )  \otimes w_m (\lam )   \right]   \right] 
 + \frac{1}{M^2}    \sum_{m \not = m'}   \mbe\left[   \tr\left[  w_m (\lam )  \otimes   w_{m'} (\lam ) \right]  \right]   \;.
\end{align}
We aim to find a lower for the above expression. Since 
\[   \sum_{m=1}^M   \mbe\left[   \tr\left[  w_m (\lam )  \otimes w_m (\lam )   \right]   \right] \succeq 0   \]
we proceed to lower bound the second term in \eqref{eq:boring}\;. Setting 
\[  \bB_{m, m'} := \Pi_m (\lam) \circ \bH \circ \Pi_{m'} (\lam) \]
for $m, m' \in [M]$ we may write 
\begin{align}
\label{eq:boring2}
 \bias (\olsest_n (\lam ))  &\geq \frac{1}{M^2}    \sum_{m \not = m'}   \mbe\left[   \tr\left[  w_m (\lam )  \otimes   w_{m'} (\lam ) \right]  \right]  \no \\
 &= \frac{1}{M^2}    \sum_{m \not = m'}   \mbe\left[     \inner{ \bH \circ \Pi_m (\lam) w^* , \Pi_{m'}(\lam)  }          \right]  \no \\ 
 &=  \frac{1}{M^2}  \sum_{m \not = m'}    \mbe\left[ \inner{  \bB_{m, m'} w^* , w^* }    \right]  \no \\
&= \frac{1}{M^2}  \sum_{m \not = m'} \left(  \sum_i \mbe\left[ (\bB_{m, m'})_{ii}  \right] (w^*_i)^2    
+  2\sum_{i>j} \mbe\left[ (\bB_{m, m'})_{ij}  \right] w^*_i \cdot w^*_j      \right)\;.
\end{align}
We now apply  Lemma \ref{lem:low-bias-1} and follow the lines of the proof of Theorem C.8 in \cite{zou2021benefits} to obtain for every $k$ 
\begin{align}
\label{eq:boring3}
\bias (\olsest_n (\lam ))  &\geq  \frac{1}{M^2}  \sum_{m \not = m'}  \sum_i \mbe\left[ (\bB_{m, m'})_{ii}  \right] (w^*_i)^2 \no \\ 
&\geq  \frac{1}{c M^2}  \sum_{m \not = m'}  \sum_i    \frac{\lam_i \cdot (w^*_i)^2 }{\left( 1 + \frac{\lam_i}{\lam_{k+1}}\cdot \frac{n}{M \rho_k} \right)^2}    \no \\  
&= \frac{M-1}{c M}  \sum_i    \frac{\lam_i \cdot (w^*_i)^2 }{\left( 1 + \frac{\lam_i}{\lam_{k+1}}\cdot \frac{n}{M \rho_k} \right)^2}    \no \\  
&\geq  \frac{M-1}{c M} \cdot \left(   \frac{M^2 \left( \lam + \sum_{j > k^*_{\ols}} \lam_j \right)^2}{n^2} \cdot ||w^*||^2_{\bH^{-1}_{0:k^*_{\ols}}}  + 
||w^*||^2_{\bH_{k^*_{\ols} : \infty}}   \right) \;, 
\end{align}
for some $c>1$.  
%and where 
%\[ \rho_k = \frac{\lam + \sum_{j > k} \lam_j}{\lam_{k+1}} \;. \] 
\end{proof}

%%%%%%%%%%%%%%%%%%%%%%%%%%%%%%%%%%%%%%%%%%%%%%%%%%%%%%%%%%%%%%%%%%%%%%%%%%%%%%%%%%%
%%%%%%%%%%%% Variance lower Bound RR
%%%%%%%%%%%%%%%%%%%%%%%%%%%%%%%%%%%%%%%%%%%%%%%%%%%%%%%%%%%%%%%%%%%%%%%%%%%%%%%%%%%

\subsubsection{Lower Bound of Variance for DRR}

\begin{proposition}[Lower Bound of Bias for local RR]
\label{prop:low-var-RR}
Suppose  $k^*_{\ols} < \frac{n}{c'M}$, for some universal constant $c'>1$. There exist constants $b,c >1$ such that 
\[  \var(\olsest_n  (\lam ))  \geq  \frac{\sigma^2}{c} \left( \frac{k^*_{\ols}}{n}  + 
\frac{n}{M^2\cdot ( \lam + \sum_{j >k^*_{\ols} } \lam_j)} \cdot \sum_{j >k^*_{\ols} } \lam_j^2\right) \;,\]
where $k^*_{\ols}$ is defined in \eqref{eq:eff-dim-RR}.
\end{proposition}

\vspace{0.2cm}

\begin{proof}[Proof of Proposition \ref{prop:low-var-RR}]
By definition of the variance, we may write
\begin{align*}
\var(\olsest_n  (\lam )) 
&=  \mbe\left[ \widehat{\var} (\olsest_n (\lam ))  \right]  \\
&=   \mbe\left[ \left| \left| \bH^{1/2} \left( \frac{1}{M} \sum_{m=1}^M ( \bX^T_m \bX_m + \lam )^{-1}\bX_m^T  \epsilon_m   \right)     \right| \right| ^2 \right] \\
&= \frac{1}{M^2} \sum_{m , m' = 1}^M \mbe_{\bX} \left[ \tr  \left[  
\bH^{1/2}    ( \bX^T_m \bX_m + \lam )^{-1}\bX_m^T  \mbe_{\epsilon_m}  \left[ \epsilon_m   \otimes \epsilon_{m'}  \right] \bX_m ( \bX^T_{m'} \bX_{m'} + \lam )^{-1}   \bH^{1/2}   \right] \right]\\
&= \frac{\sigma^2}{M^2} \sum_{m = 1}^M \mbe_{\bX} \left[\tr  \left[  
\bH^{1/2}    ( \bX^T_m \bX_m + \lam )^{-1}\bX_m^T   \bX_m ( \bX^T_m \bX_m + \lam )^{-1}   \bH^{1/2}   \right] \right] \\
&= \frac{1}{M^2} \sum_{m = 1}^M  \var ( \hat w_m^{\ols }(\lam )) \;,
\end{align*}
where the local variance is given by 
\[   \var ( \hat w_m^{\ols }(\lam ))   = \sigma^2 \mbe_{\bX} \left[\tr  \left[  
\bH^{1/2}    ( \bX^T_m \bX_m + \lam )^{-1}\bX_m^T   \bX_m ( \bX^T_m \bX_m + \lam )^{-1}   \bH^{1/2}   \right] \right] \;. \]
To lower bound the variance we utilize Theorem C.5 from \cite{zou2021benefits} (see also \cite{bartlett2020benign}) and obtain 
\begin{align*}
\var(\olsest_n  (\lam )) 
&\geq \frac{\sigma^2}{cM^2} \sum_{m = 1}^M \left( \frac{M\cdot k^*_{\ols}}{n}  + 
\frac{n}{M\cdot ( \lam + \sum_{j >k^*_{\ols} } \lam_j)} \cdot \sum_{j >k^*_{\ols} } \lam_j^2\right) \\
&= \frac{\sigma^2}{c} \left( \frac{ k^*_{\ols}}{n}  + 
\frac{n}{M^2\cdot ( \lam + \sum_{j >k^*_{\ols} } \lam_j)} \cdot \sum_{j >k^*_{\ols} } \lam_j^2\right) \;, 
\end{align*}
provided $k^*_{\ols} < \frac{n}{c'M}$, for some universal constants $c,c'>1$. 
\end{proof}

%%%%%%%%%%%%%%%%%%%%%%%%%%%%%%%%%%%%%%%%%%%%%%%%%%%%%%%%%%%%%%%%%%%%%%%%%%%%%%%%%%%%%%%%%%%%%%%%%%%%%%%%
%%%%%%%%%%%%%% Prof lower bound DRR
%%%%%%%%%%%%%%%%%%%%%%%%%%%%%%%%%%%%%%%%%%%%%%%%%%%%%%%%%%%%%%%%%%%%%%%%%%%%%%%%%%%%%%%%%%%%%%%%%%%%%%%%

\subsubsection{Proof of Theorem \ref{theo:lower-RR}}
%\vspace{0.2cm}

The proof follows by combining Proposition \ref{prop:low-bias-RR} and Proposition \ref{prop:low-var-RR} with Lemma \ref{prop:risk-local-RR}.

%%%%%%%%%%%%%%%%%%%%%%%%%%%%%%%%%%%%%%%%%%%%%%%%%%%%%%%%%%%%%%%%%%%%%%%%%%%%%%%%%%%
%%%%%%%%%%%% Upper Bound Excess Risk Tail-Averaged DSGD
%%%%%%%%%%%%%%%%%%%%%%%%%%%%%%%%%%%%%%%%%%%%%%%%%%%%%%%%%%%%%%%%%%%%%%%%%%%%%%%%%%%

\subsection{Upper Bound Excess Risk Tail-Averaged DSGD}
\label{sec:proof-upper-bound-tail-ave}

%Lemma 6.1 in \cite{zou2021benefits}

\begin{theorem}[Upper Bound Tail-averaged DSGD]
\label{theo:upp-tail-DSGD-2}
Suppose Assumption \ref{ass:well}  is satisfied. Let $\w_{M_{n}}$ denote the tail-averaged distributed estimator with $n$ training samples and assume $\gamma < 1/\tr[H]$. 
For arbitrary $k_1, k_2 \in [d]$ 
\begin{align*}
 \mbe\left[ L(\w_{M} ) \right] - L(w^*)  \; &=  \bias (\w_M) +   \var (\w_M) \; 
\end{align*}
with 
\[  \bias (\w_M) \leq \frac{c_b M^2}{\gamma^2 n^2} \cdot \left| \left|   \exp\left( -\frac{n}{M}\gamma \bH  \right) w^*\right| \right|^2_{\bH^{-1}_{0:k_1}}   +   
 || w^*||^2_{\bH_{k_1:\infty}}  \;, \]
\[ \var (\w_M) \leq   c_v (1+R^2)\cdot \sigma^2   \left( \frac{k_2 }{n}  +  \frac{n \gamma^2}{M^2} \cdot \sum_{j > k_2}\lam_j^2  \right)     \;, \]
for some universal constants $c_b , c_v>0$.
\end{theorem}

\vspace{0.2cm}

\begin{proof}[Proof of Theorem \ref{theo:upp-tail-DSGD-2}]
Utilizing \eqref{eq:bias-upp-uniform} and Lemma 6.1 in \cite{zou2021benefits}, we have 
\begin{align*}
 \bias (\w_M)  &=\frac{1}{M}\sum_{m=1}^M\bias \left(\bar w^{(m)}_{\frac{n}{M}}\right)  \\
&\leq \frac{1}{M}\sum_{m=1}^M \frac{c_b M^2}{\gamma^2 n^2} \cdot \left| \left|   \exp\left( -\frac{n}{M}\gamma \bH  \right) \right| \right|^2_{\bH^{-1}_{0:k_1}}   +   
 || w^*||^2_{\bH_{k_1:\infty}} \\
&=  \frac{c_b M^2}{\gamma^2 n^2} \cdot \left| \left|   \exp\left( -\frac{n}{M}\gamma \bH  \right) w^*\right| \right|^2_{\bH^{-1}_{0:k_1}}   +   
 || w^*||^2_{\bH_{k_1:\infty}} \;, 
\end{align*}
for some universal constant $c_b>0$.

For the variance, we utilize \eqref{eq:var-upper-uniform} and Lemma 6.1 in \cite{zou2021benefits} once more to obtain 
\begin{align*}
 \var (\w_M)  &\leq  \frac{1}{M^2}\sum_{m=1}^{M} \var \left(\bar w^{(m)}_{\frac{n}{M}}\right)  \\ 
& \leq c_v\frac{(1+R^2)\cdot \sigma^2}{M^2}\sum_{m=1}^{M}  \left( \frac{k_2 M}{n}  +  \frac{n \gamma^2}{M} \cdot \sum_{j > k_2}\lam_j^2  \right)  \\
&=  c_v (1+R^2)\cdot \sigma^2   \left( \frac{k_2 }{n}  +  \frac{n \gamma^2}{M^2} \cdot \sum_{j > k_2}\lam_j^2  \right)     \;,
\end{align*}
for some universal constant $c_v>0$.
\end{proof}

%%%%%%%%%%%%%%%%%%%%%%%%%%%%%%%%%%%%%%%%%%%%%%%%%%%%%%%%%%%%%%
%%%%%%%%%%%%% Comparing DSGD with DRR
%%%%%%%%%%%%%%%%%%%%%%%%%%%%%%%%%%%%%%%%%%%%%%%%%%%%%%%%%%%%%%

\subsection{Comparing DSGD with DRR}
\label{sec:comparison}

\begin{proof}[Proof of Theorem \ref{theo:comparison}]
To prove Theorem \ref{theo:comparison} we derive conditions on $n_\ols$ and $n_\sgd$ such that the upper bound 
for the excess risk of $\w_{M}$ for DSGD from Theorem \ref{theo:upp-tail-DSGD} can be upper bounded by the lower bound of $\olsest_n (\lam )$ for DRR 
from Theorem \ref{theo:lower-RR}, i.e. such that 
\begin{align}
\label{eq:comp-bias}
& \frac{c_b M^2}{\gamma^2 n_\sgd^2} \cdot \left| \left| \exp\left( -\frac{n_\sgd}{M}\gamma \bH  \right) w^*\right| \right|^2_{\bH^{-1}_{0:k^*_{\ols}}} +   
 || w^*||^2_{\bH_{k^*_{\ols}:\infty}} \no \\
 &\leq   \frac{M^2 \left( \lam + \sum_{j > k^*_{\ols}} \lam_j \right)^2}{c n_\ols^2} \cdot ||w^*||^2_{\bH^{-1}_{0:k^*_{\ols}}}    + ||w^*||^2_{\bH_{k^*_{\ols} : \infty}}
\end{align}
and 
\begin{align}
\label{eq:comp-var}
 c_v \left( 1+\frac{||w^*||^2}{\sigma^2} \right) \cdot \sigma^2   \left( \frac{k^*_{\ols} }{n_\sgd} + \frac{n_\sgd \gamma^2}{M^2} \cdot \sum_{j > k^*_{\ols}}\lam_j^2  \right)  
&\leq \frac{\sigma^2}{c} \left( \frac{k^*_{\ols}}{n_\ols}  + \frac{n_\ols}{M^2} \cdot
\frac{ \sum_{j >k^*_{\ols} } \lam_j^2}{ ( \lam + \sum_{j >k^*_{\ols} } \lam_j)^2} \right) \;.
\end{align}

%We start with \eqref{eq:comp-bias}. 

We start with \eqref{eq:comp-var}. For 
\[ c_v \left( 1+\frac{||w^*||^2}{\sigma^2} \right)\cdot \sigma^2   \frac{k^*_{\ols} }{n_\sgd}  \leq  \frac{\sigma^2}{c} \frac{k^*_{\ols}}{n_\ols} \]
to hold we need that 
\begin{equation}
\label{eq:cond1}
 C^* n_\ols \leq n_\sgd \;, \quad C^*:= c_v\cdot c \cdot \left( 1+\frac{||w^*||^2}{\sigma^2} \right) \;. 
\end{equation} 
 
To 
\[   c_v \left( 1+\frac{||w^*||^2}{\sigma^2} \right)\cdot \sigma^2  \frac{n_\sgd \gamma^2}{M^2} \cdot \sum_{j > k^*_{\ols}}\lam_j^2  \leq \frac{\sigma^2}{c} \frac{n_\ols}{M^2} \cdot
\frac{ \sum_{j >k^*_{\ols} } \lam_j^2}{ ( \lam + \sum_{j >k^*_{\ols} } \lam_j)^2}  \]
to hold we need
\begin{equation}
\label{eq:cond2}
 n_\sgd \leq \frac{n_\ols}{ C^* \cdot (C_\lam^*)^2 \gamma^2} \;, \quad C_\lam^* :=  \lam + \sum_{j >k^*_{\ols} } \lam_j \;. 
\end{equation} 

Finally, from \eqref{eq:comp-bias} we need 
\begin{equation}
\label{eq:cond3}
  \frac{c_b M^2}{\gamma^2 n_\sgd^2} \cdot \left| \left| \exp\left( -\frac{n_\sgd}{M}\gamma \bH  \right) w^*\right| \right|^2_{\bH^{-1}_{0:k^*_{\ols}}} 
 \leq   \frac{M^2 \left( C_\lam^* \right)^2}{c n_\ols^2} \cdot ||w^*||^2_{\bH^{-1}_{0:k^*_{\ols}}}   \;. 
\end{equation} 
 
To ensure this, note that 
\[ \left| \left| \exp\left( -\frac{n_\sgd}{M}\gamma \bH  \right) w^*\right| \right|^2_{\bH^{-1}_{0:k^*_{\ols}}} 
\leq e^{ -\frac{n_\sgd}{M}\gamma \lam_{k^*_{\ols}} } \cdot ||w^*||^2_{\bH^{-1}_{0:k^*_{\ols}}}  
\leq (1-\gamma \lam_{k^*_{\ols}} )\cdot ||w^*||^2_{\bH^{-1}_{0:k^*_{\ols}}}  \;. \]
Hence, \eqref{eq:cond3} is implied if 
\begin{equation*}
 \frac{c_b}{\gamma^2 n_\sgd^2} (1-\gamma \lam_{k^*_{\ols}} ) \leq  \frac{\left( C_\lam^* \right)^2}{c n_\ols^2} \;, 
\end{equation*} 
being equivalent to 
\begin{equation}
\label{eq:cond4}
 \frac{\sqrt{cc_b( 1-\gamma \lam_{k^*_{\ols}} )}}{\gamma C_\lam^*} n_\ols \leq  n_{\sgd} \;.
\end{equation} 
Combining conditions \eqref{eq:cond1}, \eqref{eq:cond2} and \eqref{eq:cond4} we need 
\[ \max \left\{ C^* , \frac{\sqrt{cc_b( 1-\gamma \lam_{k^*_{\ols}} )}}{\gamma C_\lam^*}  \right\}\cdot  n_\ols 
\leq n_\sgd 
\leq  \frac{1}{ C^* \cdot (C_\lam^*)^2 \gamma^2}  \cdot n_\ols   \;. \] 
A short calculation shows that the condition 
\[ \gamma < \min\left\{ \frac{1}{\tr[H]} , \frac{1}{\sqrt{c}C^* C_\lam^*} \right\} \]
implies that 
\[  \max \left\{ C^* , \frac{\sqrt{cc_b( 1-\gamma \lam_{k^*_{\ols}} )}}{\gamma C_\lam^*}  \right\} \leq   \frac{1}{ C^* \cdot (C_\lam^*)^2 \gamma^2}  \;. \]
This finishes the proof. 
\end{proof}

%%%%%%%%%%%%%%%%%%%%%%%%%%%%%%%%%%%%%%%%%%%%%%%%%%%%%%%%%%%%%%
%%%%%%%%%%%%% FURTHER EXPERIMENTS
%%%%%%%%%%%%%%%%%%%%%%%%%%%%%%%%%%%%%%%%%%%%%%%%%%%%%%%%%%%%%%

\section{FURTHER NUMERICAL EXPERIMENTS}
\label{sec:further-numerics}

In this Section we collect further experimental results conducted on simulated data from Section \ref{sec:numerics}.

\begin{figure}[h]
\centering
\includegraphics[width=0.4\columnwidth, height=0.25\textheight]{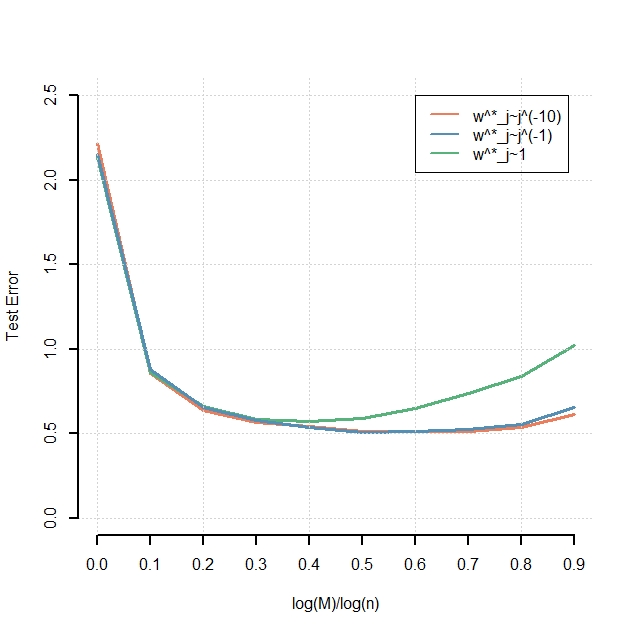}
\caption{Test error for distributed ridgeless regression with $\lam_j = j^{-2}$ for different sources $w^*$ as a function of $M=n^\alpha$, 
$\alpha \in \{0, 0.1,...,0.9\}$. The number of local nodes acts as a regularization parameter. We generate $n=500$ i.i.d. 
training data with $x_j \sim \cN(0, \bH)$ with 
mildly overparameterization $d=700$.}
\label{fig:3}
\end{figure}

\begin{figure}[h]
\centering
\includegraphics[width=0.3\columnwidth, height=0.23\textheight]{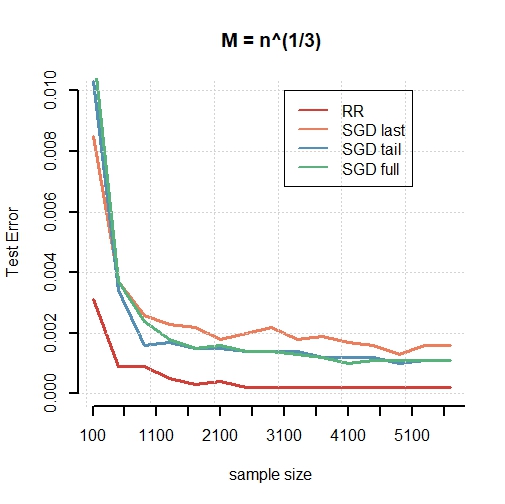}
\includegraphics[width=0.3\columnwidth, height=0.23\textheight]{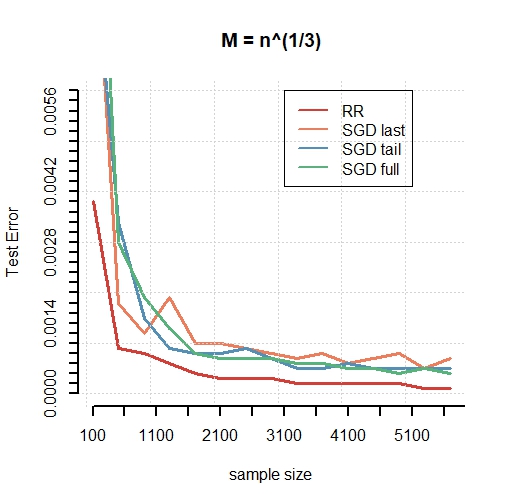}
\includegraphics[width=0.3\columnwidth, height=0.23\textheight]{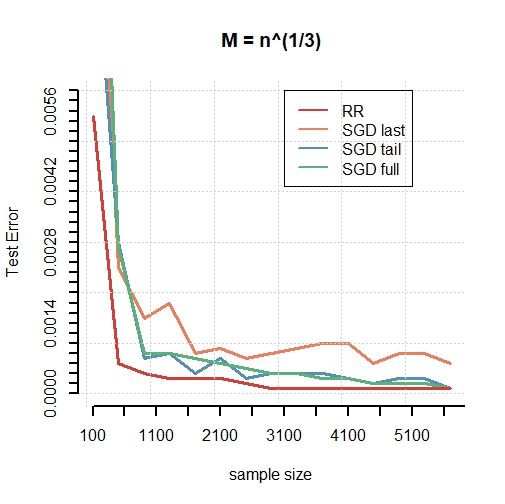}
\caption{{\bf Left:} $\lam_j = j^{-10}$, $w^*_j = 1$ {\bf Middle:} $\lam_j = j^{-10}$, $w^*_j = j^{-1}$ {\bf Right:} $\lam_j = j^{-10}$, $w^*_j = j^{-10}$ }
\label{fig:6}
\end{figure}

\begin{figure}[h]
\centering
\includegraphics[width=0.3\columnwidth, height=0.23\textheight]{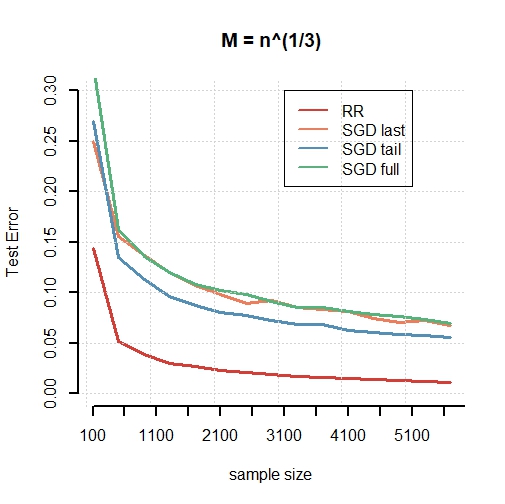}
\includegraphics[width=0.3\columnwidth, height=0.23\textheight]{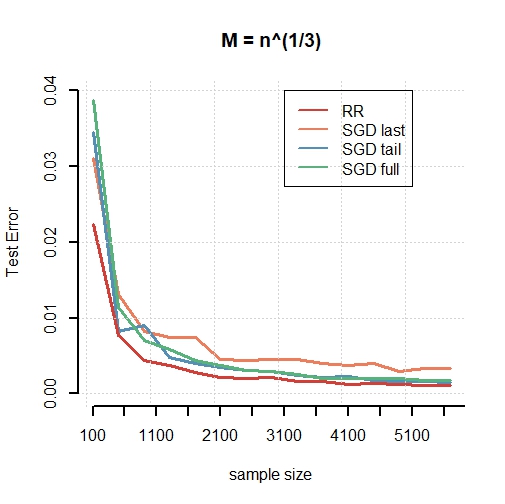}
\includegraphics[width=0.3\columnwidth, height=0.23\textheight]{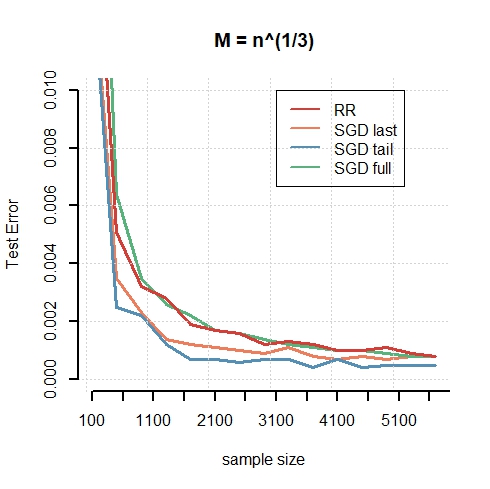}
\caption{{\bf Left:} $\lam_j = j^{-2}$, $w^*_j = 1$ {\bf Middle:} $\lam_j = j^{-2}$, $w^*_j = j^{-1}$ {\bf Right:} $\lam_j = j^{-2}$, $w^*_j = j^{-10}$ }
\label{fig:7}
\end{figure}

We compare the sample complexity of optimally tuned full-averaged DSGD, tail-averaged DSGD and last-iterate DSGD with optimally tuned DRR for different sources $w^*$, 
see Figures \ref{fig:6}, \ref{fig:7} and \ref{fig:7}. Here, the data are generated as in Section \ref{sec:numerics}
with $d=200$, $\lam_j = j^{-2}$ and $w^*_j=j^{-\alpha}$, $\alpha \in \{0, 1, 10\}$. 
The number of local nodes is fixed at $M_n=n^{1/3}$ for each $n \in \{100, ..., 6000\}$. 
%For this problem instance, DSGD may perform even better than DRR for sparse targets ($\alpha=10$), 
%i.e.,  DSGD achieves the same accuracy as DRR with less samples in this regime. For less sparse targets $\alpha=1$, the sample complexities of DSGD and DRR 
%are comparable while for non-sparse targets ($\alpha=0$), DRR outperforms DSGD.     

\begin{figure}[h]
\centering
\includegraphics[width=0.3\columnwidth, height=0.23\textheight]{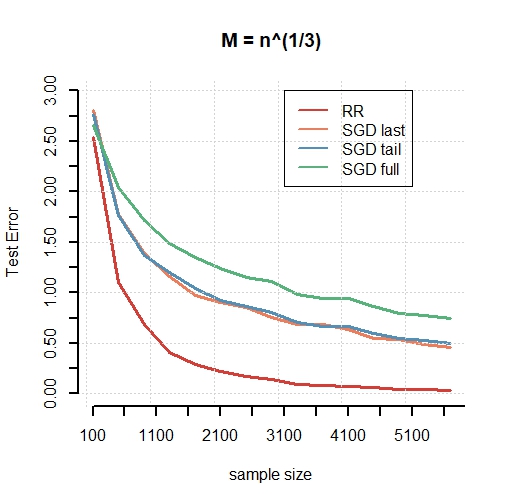}
\includegraphics[width=0.3\columnwidth, height=0.23\textheight]{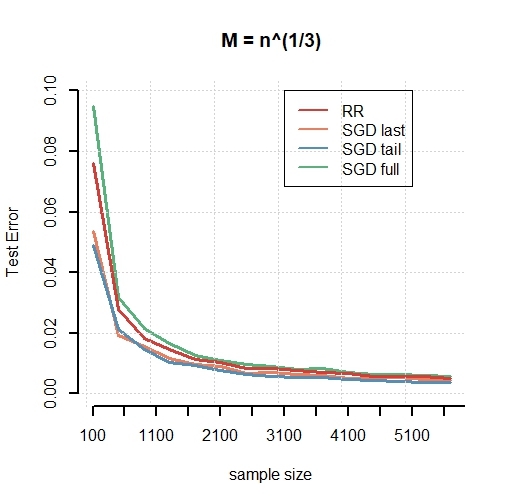}
\includegraphics[width=0.3\columnwidth, height=0.23\textheight]{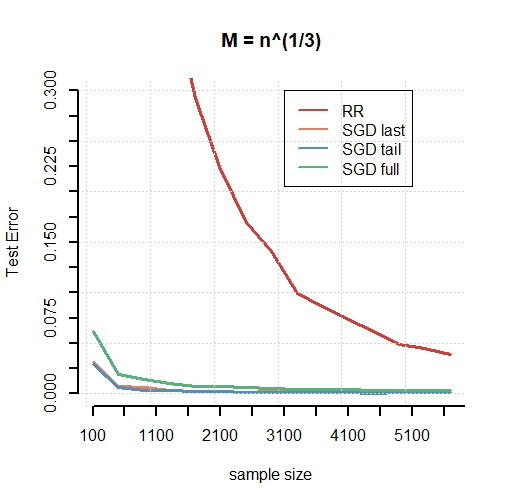}
\caption{{\bf Left:} $\lam_j = j^{-1}$, $w^*_j = 1$ {\bf Middle:} $\lam_j = j^{-1}$, $w^*_j = j^{-1}$ {\bf Right:} $\lam_j = j^{-1}$, $w^*_j = j^{-10}$ }
\label{fig:8}
\end{figure}

%%%%%%%%%%%%%%%%%%%%%%%%%%%%%%%%%%%%%%%%%%%%%%%%%%%%%%%%%%%%%%%%%%%%%%%%%%%%%%%%%%%%%%%%%%%%%%%%%%%%%%%%%%%
%%%%%%%%%%%%%%%%%% (A) PROOFS MAIN RESULTS  
%%%%%%%%%%%%%%%%%%%%%%%%%%%%%%%%%%%%%%%%%%%%%%%%%%%%%%%%%%%%%%%%%%%%%%%%%%%%%%%%%%%%%%%%%%%%%%%%%%%%%%%%%%%

\end{document}